\documentclass[3p,times]{elsarticle}
    \usepackage{tikz}
\usetikzlibrary{arrows.meta, positioning}
\usetikzlibrary{positioning, arrows.meta, fit}
\usepackage{amsmath, amssymb, bm}          
\usepackage{tikz}
\usetikzlibrary{
  calc,                
  positioning,        
  decorations.pathreplacing, 
  arrows.meta         
}
\usepackage{booktabs}
\usepackage{siunitx}
\usepackage{amsmath}
\definecolor{answerblue}{RGB}{0,51,102}
    \usepackage{hyperref}
    \usepackage{booktabs}
    \usepackage{graphicx}
    \usepackage{placeins}
    \usepackage{lipsum}
    \usepackage{tikz}
\usetikzlibrary{positioning, arrows.meta, shapes.geometric}
\usepackage[table]{xcolor}      
\usepackage{colortbl}            
\usepackage{booktabs}           
\usepackage{multirow}            
\usepackage{siunitx}             
    \usepackage{xcolor}
\usepackage[utf8]{inputenc}
\usepackage[T1]{fontenc}
\usepackage{amsmath,amsfonts,amssymb} 
\usepackage{enumitem}                
\usepackage{caption}                 
\usepackage{tikz}
\usetikzlibrary{positioning,decorations.pathreplacing,arrows.meta,fit}
\usepackage[table,xcdraw]{xcolor} 
    \usepackage{colortbl} 
    \usepackage[ruled,vlined]{algorithm2e}   
    \usepackage{textcomp}
    \usepackage{algpseudocode}
    \usepackage[utf8]{inputenc}
    \usepackage[T1]{fontenc}
    \usepackage{bm}
    \usepackage{subcaption}
    \usepackage{multirow}
    \usepackage{array}
    \usepackage{multicol}
    \usepackage{comment}
    \usepackage{color,soul}
    \usepackage{amsmath}
    \usepackage{natbib}
\usepackage{graphicx}
\usepackage{subcaption}

\usepackage[capitalize]{cleveref}
\Crefformat{equation}{#2Eq.~(#1)#3}

\usepackage{amsthm}

\usepackage{float} 

\usepackage{graphicx,tikz}
\graphicspath{{./img/}}
\usetikzlibrary{bayesnet}

    \journal{Journal Name}

    \title{Amortized Variational Inference for Logistic Regression with Missing Covariates}
    
    \author[1]{M. Cherifi}
\ead{cherifi.meddd@gmail.com}
\author[2]{A. Sportisse}
\ead{Aude.Sportisse@univ-grenoble-alpes.fr}

\author[3]{X. Zhu}
\ead{xujia.zhu@centralesupelec.fr}
\author[3]{M. N. El Korso}
\ead{mohammed.nabil.el-korso@centralesupelec.fr}

\author[1]{A. Mesloub}
\ead{meslouba@gmail.com}

\cortext[cor1]{Corresponding author}
\address[1]{Lab. Traitement du signal, Ecole Militaire Polytechnique, BP 17 Bordj El Bahri, Algeria}
\address[2]{Lab. d'Informatique de Grenoble, CNRS, France }
\address[3]{University of Paris Saclay, CNRS, CentraleSupelec, Laboratoire des signaux et systèmes, 91190, Gif-sur-Yvette, France}

\begin{document}

\begin{abstract}
Missing covariate data pose a significant challenge to statistical inference and machine learning, particularly for classification tasks, where models like logistic regression are widely used. For the latter, classical iterative approaches, such as the Expectation--Maximization (EM) algorithm and its stochastic variants, or multiple imputation methods, are often computationally intensive, sensitive to high rates of missingness, and limited in their ability to propagate uncertainty. Recent deep generative models based on Variational AutoEncoders (VAEs) have shown promise in handling missing data through amortized inference, but they often rely on complex latent representations.

In this manuscript, we propose Amortized Variational Inference for Logistic Regression (AV-LR), a unified, end-to-end framework for binary logistic regression with missing covariates. AV-LR integrates a probabilistic generative model with a simple amortized inference network, trained jointly by maximizing the evidence lower bound. Unlike competing methods, AV-LR directly performs inference in the space of missing data without introducing additional latent variables, leveraging a single inference network and a linear layer that jointly estimate the regression parameters and the parameters of the missingness mechanism.

We demonstrate that AV-LR achieves estimation accuracy comparable to or better than state-of-the-art EM-like algorithms, while significantly reducing computational cost through a simple deep learning-based architecture, in contrast to more complex VAE-based approaches. Furthermore, AV-LR naturally extends to missing-not-at-random settings, where the missing data depend on both the observed and missing values, by explicitly modeling the missingness mechanism. Empirical results on synthetic and real-world datasets confirm the effectiveness and efficiency of AV-LR across various missing-data scenarios.
\end{abstract}

\maketitle

\section{Introduction}

Missing data present a substantial challenge in statistical inference and machine learning, often introducing bias and reducing predictive power when not properly addressed \cite{hippert2025missing,hippert2025missing01}. 
Traditional approaches based on the Expectation--Maximization (EM) algorithm \cite{dempster1977maximum} and its stochastic variants, such as Stochastic EM (SEM) \cite{wei1990monte} and Stochastic Approximation EM (SAEM) \cite{kuhn2005maximum,jiang2020logistic,cherifi2025maximum}, mitigate the intractability of directly evaluating the observation likelihood by iteratively imputing or simulating missing values. However, these iterative methods can be computationally intensive and sensitive to both data dimensionality and missing rates, especially in high-dimensional logistic regression models \cite{cherifi2025robust,jiang2020logistic}. {Multiple imputation techniques, such as Multiple Imputation by Chained Equations (MICE) \cite{van2011mice}, and iterative random forest-based methods like missForest \cite{stekhoven2012missforest} provide flexibility by modeling each variable conditionally. 
More specifically, these approaches iteratively treat each incomplete variable as a response and predict its missing values using the other variables as predictors. 
Nevertheless, they become computationally demanding on large datasets, and do not fully leverage a joint modeling framework for the covariate distribution and the missing-data mechanism, which may limit the coherence of subsequent inference. Moreover, these classical methods are not designed to handle Missing Not At Random (MNAR) data, i.e., when the missingness depends on the missing values themselves. A classic example of MNAR data occurs when individuals with very high or very low incomes are less likely to report them due to their value. } 

Recent methods using deep generative modeling, particularly Variational AutoEncoders (VAEs) \cite{kingma2013auto} and their Importance-Weighted extensions (IWAEs) \cite{burda2015importance}, have introduced amortized inference networks to efficiently approximate posterior distributions. 
While these VAE/IWAE-based approaches, e.g., MIWAE \cite{mattei2019miwae}, not-MIWAE for MNAR data \cite{ipsen2020not} and related VAE imputations \cite{lim2024deeply,gong2021variational,ipsen2020not,collier2020vaes}, offer powerful reconstruction abilities, they present several practical limitations that motivate our work. 
Many methods introduce additional latent spaces and multiple neural components, thereby increasing model complexity, complicating optimization and hyperparameter tuning, and reducing interpretability. 
Second, existing approaches typically focus on likelihood or reconstruction objectives and do not jointly optimize the imputation process with the downstream predictive model and the missingness mechanism, which can degrade inferential validity for tasks such as logistic regression, especially under MNAR. 
A recent study \cite{lim2024deeply} extends the not-MIWAE framework \cite{ipsen2020not} to generalized linear models with MNAR covariates by introducing a latent VAE representation and optimizing an importance-weighted autoencoder objective to tighten the bound on the marginal likelihood under non-ignorable missingness. 

These shortcomings motivate this work, where we develop an end-to-end, computationally efficient scheme. This framework performs amortized inference directly in the space of missing covariates, avoids introducing extra latent variables, and couples a single, simple inference network with a single linear component that jointly models prediction and missingness, thereby improving interpretability, stability, and task-oriented coherence. More specifically, we propose in this work a unified framework for binary logistic regression with missing covariates, referred to as Amortized Variational Inference for Logistic Regression (AV-LR). Building on recent theoretical foundations for amortized variational inference \cite{margossian2023amortized,ganguly2023amortized}, AV-LR combines a probabilistic generative model that defines the joint likelihood $p(\mathbf{y}, \mathbf{x})$ with an amortized inference network $q_{\bm{\phi}}(\mathbf{x}_{\mathrm{mis}} \mid \mathbf{x}_{\mathrm{obs}}, y)$, trained jointly by maximizing the ELBO \cite{kingma2013auto,kim2018semi}. Via numerical examples, we show that AV-LR achieves comparable estimation accuracy to SEM/SAEM while substantially reducing computational cost, even under high missingness rates and non‑ignorable missingness mechanisms (MNAR). 

The remainder of this article is structured as follows. \Cref{SectionMS_binary} introduces the probabilistic logistic regression model with missing data and reviews the foundational aspects of the SAEM algorithm. \Cref{SectionAVLR} details the AV-LR methodology, including the specification of the generative model, the architecture of the inference network, and the derivation of both the standard and importance-weighted ELBO. \Cref{SectionMNAR} extends AV-LR to MNAR settings through explicit modeling of the missingness mechanism. Section \ref{sec:prediction} focuses how to predict an output variable using the trained model. Section \ref{sec:comparison_VAE} presents a detailed comparison of AV-LR methodology and VAE-based methods. Finally,  \Cref{SectionExperiments} provides empirical results on synthetic and real‑world datasets, comparing AV‑LR against competing approaches in terms of estimation accuracy and computational efficiency.

\section{Model Setup} 
\label{SectionMS_binary}

We consider a conventional classification problem: we are given \(N\) independent observations $\{(\boldsymbol{x}_i,y_i):i=1,\ldots,N\}$, where each label $y_i \in \{0,1\}$ conditional on covariates \(\boldsymbol{x}_i \in \mathbb{R}^d\) is modeled as a realization from a Bernoulli distribution. Specifically, the probability that the binary response \(y_i\) equals 1 is modeled via the logistic function:

\begin{equation}
    \label{LR_model_binary}
    p(y_i = 1 \mid \boldsymbol{x}_i; \boldsymbol{\beta}) 
    = \frac{\exp\!\bigl([1,\;\boldsymbol{x}_i^\top]\,\boldsymbol{\beta}\bigr)}{1 + \exp\!\bigl([1,\;\boldsymbol{x}_i^\top]\,\boldsymbol{\beta}\bigr)},
    \quad
    p(y_i = 0 \mid \boldsymbol{x}_i; \boldsymbol{\beta}) 
    = 1 - p(y_i = 1 \mid \boldsymbol{x}_i; \boldsymbol{\beta}),
\end{equation}

where \(\boldsymbol{\beta} \in \mathbb{R}^{d+1}\) is the unknown parameter vector (including the intercept). 

In practice, covariate data are sometimes incomplete, with certain variables unobserved. To explicitly account for the missingness, we partition the covariate vector as \(\boldsymbol{x}_i = (\boldsymbol{x}_i^{\mathrm{obs}},\,\boldsymbol{x}_i^{\mathrm{mis}})\), distinguishing observed components $\boldsymbol{x}_i^{\mathrm{obs}}$ from missing components $\boldsymbol{x}_i^{\mathrm{mis}}$. In addition, we define a binary \emph{mask vector} $\bm{r}_i \in \{0,1\}^p$ associated with each observation $\bm{x}_i$. The mask vector encodes the missingness pattern of the covariates, such that
\[
  r_{ij} =
  \begin{cases}
    1, & \text{if the $j$-th component of $\bm{x}_{i}$ is observed}, \\[6pt]
    0, & \text{if the $j$-th component of $\bm{x}_{i}$ is missing}.
  \end{cases}
\]

According to the underlying missingness mechanism \cite{rubin1987statistical}, three classical cases are distinguished. 
Missing Completely At Random (MCAR) holds when the missingness pattern is independent of both observed and unobserved covariates and the response, i.e., \(\Pr(\boldsymbol{r}_i\mid\boldsymbol{x}_i,y_i)=\Pr(\boldsymbol{r}_i)\). 
Missing At Random (MAR) allows dependence on observed quantities but not on the missing values themselves: \(\Pr(\boldsymbol{r}_i\mid\boldsymbol{x}_i,y_i)=\Pr(\boldsymbol{r}_i\mid\boldsymbol{x}_i^{\mathrm{obs}},y_i)\). 
Missing Not At Random (MNAR) refers to situations where the missingness mechanism depends explicitly on both the observed and unobserved components, i.e., \(\Pr(\boldsymbol{r}_i \mid \boldsymbol{x}_i, y_i) = \Pr(\boldsymbol{r}_i \mid \boldsymbol{x}_i^{\mathrm{obs}}, \boldsymbol{x}_i^{\mathrm{mis}}, y_i)\).
 In this section, we restrict our attention to the MCAR/MAR cases and defer discussion of MNAR to Section~\ref{SectionMNAR}.

 To model the covariates probabilistically and account for different patterns of missingness, we assume the complete covariate vector \(\boldsymbol{x}_i\) follows a multivariate normal distribution \(\boldsymbol{x}_i \sim \mathcal{N}(\boldsymbol{\mu},\boldsymbol{\Sigma})\), where \(\boldsymbol{\mu}\in\mathbb{R}^d\) and \(\boldsymbol{\Sigma}\in\mathbb{R}^{d\times d}\). This choice plays a critical role in our modeling framework, as it avoids introducing complex latent variables while retaining a fully probabilistic approach. Despite its simplicity, this assumption is particularly advantageous in real-world problems with limited data. It is commonly adopted for modeling continuous multivariate variables, especially when handling missing data \cite{jiang2020logistic,cherifi2025robust,ibrahim1999missing,dempster1977maximum}, and provides a flexible approximation to a wide range of dependence structures. Following this setup, when covariates are partially missing, the observed-data likelihood is obtained by marginalizing over the missing components:
\begin{equation}\label{log_obs_binary}
\begin{split}
    \mathcal{L}(\boldsymbol{\beta}, \boldsymbol{\mu}, \boldsymbol{\Sigma}; \boldsymbol{y}, \{\boldsymbol{x}_i^{\mathrm{obs}}\}_{i=1}^N) 
    &= \prod_{i=1}^N p(y_i, \boldsymbol{x}_i^{\mathrm{obs}} \mid \boldsymbol{\beta}, \boldsymbol{\mu}, \boldsymbol{\Sigma}) \\
    &= \prod_{i=1}^N 
      \int 
        p(y_i = 1 \mid \boldsymbol{x}_i; \boldsymbol{\beta})^{y_i}
        \,\bigl[1 - p(y_i = 1 \mid \boldsymbol{x}_i; \boldsymbol{\beta})\bigr]^{1 - y_i}
        \; f(\boldsymbol{x}_i; \boldsymbol{\mu}, \boldsymbol{\Sigma}) \; 
      \mathrm{d}\boldsymbol{x}_i^{\mathrm{mis}},
\end{split}
\end{equation}

where $f(\cdot;\boldsymbol{\mu},\boldsymbol{\Sigma})$ denotes the multivariate normal probability density function with mean vector $\boldsymbol{\mu}$ and covariance matrix $\boldsymbol{\Sigma}$, characterizing the joint distribution of covariates. The presence of missing covariates renders the marginal likelihood in \Cref{log_obs_binary} analytically intractable due to the required integration over the missing data space. This computational challenge motivates the adoption of a Stochastic Approximation Expectation-Maximization (SAEM) framework, which circumvents direct integration by combining Monte Carlo sampling of the missing data with stochastic approximation in the E-step.

At each iteration \(t\), the SAEM algorithm proceeds as follows:

\begin{enumerate}
  \item \textbf{Simulation Step (S-Step):}\\
    For each observation \(i\), draw a Monte Carlo sample of the missing covariates from the conditional distribution:
    \[
      \boldsymbol{x}_{i,\mathrm{mis}}^{(t+1)}
      \sim
      p\!\bigl(\boldsymbol{x}_{i,\mathrm{mis}}
           \mid \boldsymbol{y}_i,\,\boldsymbol{x}_{i,\mathrm{obs}}
           ;\,\boldsymbol{\beta}^{(t)},\,\boldsymbol{\mu}^{(t)},\,\boldsymbol{\Sigma}^{(t)}\bigr).
    \]

  \item \textbf{Stochastic Approximation (SA-Step):}\\
    Update the intermediate expectation of the complete-data log-likelihood via a stochastic approximation scheme:
    \[
      Q_{t+1} = Q_t + \gamma_t \Big( \log p(\boldsymbol{y}, \boldsymbol{x}_{\mathrm{obs}}, \boldsymbol{x}_{\mathrm{mis}}^{(t+1)};\,
      \boldsymbol{\beta}^{(t)},\,\boldsymbol{\mu}^{(t)},\,\boldsymbol{\Sigma}^{(t)}) - Q_t \Big),
    \]
    where \(\gamma_t\) is a decreasing step size sequence.

  \item \textbf{Maximization Step (M-Step):}\\
    Update parameters by maximizing the stochastic surrogate:
    \[
      (\boldsymbol{\beta}^{(t+1)},\,\boldsymbol{\mu}^{(t+1)},\,\boldsymbol{\Sigma}^{(t+1)})
      = \arg\max_{\boldsymbol{\beta},\,\boldsymbol{\mu},\,\boldsymbol{\Sigma}} \, Q_{t+1}.
    \]
\end{enumerate}
Initially, \(Q_{0}\) is computed using simple mean imputation for the missing covariates, which provides a starting point for the stochastic approximations.
The SAEM algorithm ensures stochastic convergence toward the maximum likelihood estimates, with the decreasing step size \(\gamma_t\) guaranteeing asymptotic stability. Moreover, the algorithm is particularly efficient in high-dimensional settings, where the traditional EM algorithm may be computationally prohibitive.

However, the SAEM algorithm suffers from a significant drawback: the computational cost associated with the sampling step (S-Step) can be prohibitively high, especially for large datasets or complex models \cite{cherifi2025robust}. The need to draw Monte Carlo samples from the conditional distribution of missing covariates for each observation at each iteration requires substantial computational resources and can lead to slow convergence. This limitation motivates the exploration of alternative approaches to handle the intractable integral in the likelihood of the observed data. In the next section, we present an amortized variational inference framework that addresses these computational challenges by approximating the posterior distribution of missing covariates using a neural network, thereby reducing the computational burden while maintaining statistical accuracy.

\section{AV-LR: Amortized Variational Inference for Logistic Regression with Missing Data} \label{SectionAVLR}

\begin{figure}[!htbp]
\centering
    \begin{tikzpicture}
        \node[obs] (y) {$y$} ; %
        \node[latent, above=of y, path picture={\fill[gray!25] (path picture bounding box.south) rectangle (path picture bounding box.north west);}] (x) {$x$} ; %
        \node[const, right=of x] (musig) {$\mu,\Sigma$} ;
        \node[const, right=of y](beta) {$\beta$};
        \node[const, left=of x](phi) {$\phi$};
        
        \edge {x} {y} ; %
        \edge {musig}{x};
        \edge {beta}{y};
        \edge[dashed] {phi}{x};
        
        \path (y) edge[dashed, bend right, ->] (x);
        
        \plate {yx} {(y)(x)} {$N$};
    \end{tikzpicture}
         \caption{ Structural causal graph of AV-LR for MCAR or MAR covariates. Nodes in grey represent fully observed variables, nodes in white denote unobserved variables, and mixed nodes indicate the presence of both cases. he edges from $x$ to $y$ means that $x$ causes $y$.} 
         \label{fig:graphical_MCAR}
\end{figure}
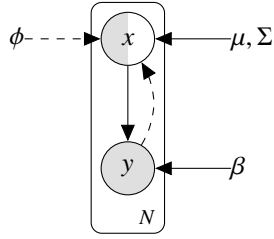

In this section, we present the proposed AV-LR framework for binary logistic regression with missing covariates. More specifically, we introduce an inference network \( q_{\bm{\phi}}(\bm{x}_{\text{miss}} \mid \bm{x}_{\text{obs}}, \bm{y}) \) parameterized by neural network weights \( \bm{\phi} \) that approximates the conditional posterior distribution of the missing covariates given the observed data and outcomes. This approach leverages amortized variational inference to directly estimate the missing values, eliminating the need for iterative imputation or sampling-based methods typical of EM-like algorithms.
This network enables efficient and scalable inference by sharing parameters across all data points, facilitating rapid approximation even in high-dimensional settings \cite{kim2018semi}. \Cref{fig:graphical_MCAR}
shows the causal relationships between the variables $x$ and $y$ through a structural causal graph, when the missingness mechanism is MCAR or MAR. 

Recall that the joint model \( p_{\bm{\theta}}(y_i, \bm{x}_i) \) combines a logistic regression likelihood and a multivariate Gaussian prior on the covariates:
\begin{align}
    p_{\bm{\theta}}(y_i \mid \bm{x}_i) &= \sigma\left(\bm{\beta}^\top \tilde{\bm{x}}_i\right)^{y_i} \left(1 - \sigma\left(\bm{\beta}^\top \tilde{\bm{x}}_i\right)\right)^{1 - y_i}, \quad \tilde{\bm{x}}_i = [1, \bm{x}_i] \in \mathbb{R}^{d+1}, \\
    p_{\bm{\theta}}(\bm{x}_i) &= f\left(\bm{x}_i ; \bm{\mu}, \bm{\Sigma}\right),
\end{align}
where \( \bm{\beta} \in \mathbb{R}^{d+1} \) includes the bias term, \( \bm{\mu} \in \mathbb{R}^d \) is the mean vector, \( \bm{\Sigma} \in \mathbb{R}^{d \times d} \) is the covariance matrix, and $\sigma$ denotes the sigmoid function given by
\[
\sigma(u) = \frac{1}{1 + e^{-u}}.
\]
The full parameters are \( \bm{\theta} = (\bm{\beta}, \bm{\mu}, \bm{\Sigma}) \).

\subsection{Variational Lower Bound and Importance-Weighted Estimation}
\label{subsec:elbo_iwelbo}

The likelihood of the observed data given in \eqref{log_obs_binary} is intractable due to the integration over missing covariates. Using Jensen's inequality, we obtain the evidence lower bound (ELBO) for the observed log-likelihood:
\begin{equation}\label{eq:ELBO}
    \log p_{\bm{\theta}}(y_i, \bm{x}_{i,\text{obs}}) \geq \mathbb{E}_{q_{\bm{\phi}}}\left[\log \frac{p_{\bm{\theta}}(y_i, \bm{x}_{i,\text{miss}}, \bm{x}_{i,\text{obs}})}{q_{\bm{\phi}}(\bm{x}_{i,\text{miss}} \mid \bm{x}_{i,\text{obs}}, y_i)}\right] \triangleq \mathcal{L}_i(\boldsymbol{\theta},\boldsymbol{\phi}),
\end{equation}
where $q_{\bm{\phi}}$ is a variational distribution parameterized by $\bm{\phi}$, introduced to approximate the true posterior distribution $p_{\bm{\theta}}(\bm{x}_{i,\text{miss}}\mid \bm{x}_{i,\text{obs}},y_i)$. Expanding the joint likelihood as \( \log p_{\bm{\theta}}(y_i, \bm{x}_i) = \log p_{\bm{\theta}}(y_i \mid \bm{x}_i) + \log p_{\bm{\theta}}(\bm{x}_i) \), the ELBO decomposes into three terms:
\[
\mathcal{L}_i(\boldsymbol{\theta},\boldsymbol{\phi})
= \mathbb{E}_{q_{\bm{\phi}}}\!\big[\log p_{\bm{\theta}}(y_i\mid \bm{x}_i)\big]
+ \mathbb{E}_{q_{\bm{\phi}}}\!\big[\log p_{\bm{\theta}}(\bm{x}_i)\big]
- \mathbb{E}_{q_{\bm{\phi}}}\!\big[\log q_{\bm{\phi}}(\bm{x}_{i,\mathrm{miss}}\mid \bm{x}_{i,\mathrm{obs}},y_i)\big].
\]

In practice, estimating gradients of the parameters \( (\bm{\theta}, \bm{\phi}) \) with a single sample from the variational posterior often leads to high variance and slow convergence. To address this, we adopt the Importance-Weighted Autoencoder (IWAE) objective, which uses multiple importance samples to tighten the bound and stabilize gradient estimates \cite{burda2015importance}.

For each observation \( i \), we draw \( K \) independent samples from the variational posterior:
\[
 \bm{x}_{i,\mathrm{miss}}^{(k)} \sim q_{\bm{\phi}}\bigl(\bm{x}_{i,\mathrm{miss}} \mid \bm{x}_{i,\mathrm{obs}}, y_i\bigr), \quad k = 1, \dots, K.
\]
Each sample is then used to reconstruct the complete covariate vector by combining observed and imputed values:
\[
  \bm{x}_i^{(k)}
  = \bm{x}_{i,\mathrm{obs}} \odot \bm{r}_i
  + \bm{x}_{i,\mathrm{miss}}^{(k)} \odot \bigl(\mathbf{1} - \bm{r}_i\bigr),
\]
where \( \bm{r}_i \) is the binary mask indicating observed entries. The log-importance weight for the \(k\)-th sample is defined as
\[
\log w_i^{(k)}
= \log p_{\bm{\theta}}\bigl(y_i \mid \bm{x}_i^{(k)}\bigr)
+ \log p_{\bm{\theta}}\bigl(\bm{x}_i^{(k)}\bigr)
- \log q_{\bm{\phi}}\bigl(\bm{x}_{i,\mathrm{miss}}^{(k)} \mid \bm{x}_{i,\mathrm{obs}}, y_i\bigr),
\]
which quantifies the log-discrepancy between the model joint density evaluated at the imputed complete covariate vector and the variational proposal distribution that generated the imputation. The corresponding unnormalized importance weights are obtained as 
$w_i^{(k)}=\frac{p_{\bm{\theta}}\bigl(y_i \mid \bm{x}_i^{(k)}\bigr) p_{\bm{\theta}}\bigl(\bm{x}_i^{(k)}\bigr)}{q_{\bm{\phi}}\bigl(\bm{x}_{i,\mathrm{miss}}^{(k)} \mid \bm{x}_{i,\mathrm{obs}}, y_i\bigr)}$.
Aggregating over the \( K \) samples yields the importance-weighted ELBO (IWELBO):
\begin{equation}\label{eq:iwelbo_improved}
 \mathcal{L}_{\mathrm{IW}}(\bm{\theta}, \bm{\phi}) = \sum_{i=1}^N\Biggl[\log \Bigl(\tfrac{1}{K} \sum_{k=1}^K  w_i^{(k)} \Bigr)\Biggr] \;\le\; \log p_{\bm{\theta}}\bigl(\bm{y}, \bm{x}_{\mathrm{obs}}\bigr).
\end{equation}
Note that increasing the number of importance samples \( K \) tightens the bound and reduces the variance of gradient estimates, leading to more stable optimization and improved parameter estimation. Finally, we train the model by minimizing the negative IWELBO, \( -\mathcal{L}_{\mathrm{IW}} \), using stochastic gradient descent.

\subsection{Amortized Variational Approximation}
\label{subsec:amortized_variational}

In our amortized variational inference scheme, we model the variational distribution \( q_{\bm{\phi}}(\bm{x}_{i,\text{miss}} \mid \bm{x}_{i,\text{obs}}, y_i) \) as a multivariate Gaussian distribution using a neural network encoder with the parameter $\bm{\phi}$. The variational distribution is explicitly given by
\begin{equation*}
    q_{\bm{\phi}}(\bm{x}_{i,\text{miss}} \mid \bm{x}_{i,\text{obs}}, y_i) 
    = \mathcal{N}\bigl( \bm{x}_{i,\text{miss}} \,\big|\, \bm{\mu}_{q,i},\, \bm{\Sigma}_{q,i} \bigr),
    \label{eq:variational_dist}
\end{equation*}
where \( \bm{\mu}_{q,i} \in \mathbb{R}^{d_i} \) is the mean vector and covariance matrix \( \bm{\Sigma}_{q,i} \) corresponds to the covariance matrix. 

Let \( d_i \) denote the number of missing covariates for observation \( i \). The encoder network takes as input the concatenated vector \( [\bm{x}_{i,\text{obs}}, y_i] \) and outputs the mean vector \( \bm{\mu}_{q,i} \in \mathbb{R}^{d_i} \) and a vector \( \bm{\ell}_{q,i} \in \mathbb{R}^{d_i(d_i+1)/2} \) that encodes the lower-triangular entries of the Cholesky factor of the covariance matrix. Then, \( \bm{\Sigma}_{q,i} \) is constructed as \( \bm{\Sigma}_{q,i} = \bm{L}_{q,i} \bm{L}_{q,i}^\top \), where \( \bm{L}_{q,i} \) is a lower-triangular matrix obtained via the mapping
\[
\bm{L}_{q,i} = \texttt{CholeskyFactor}(\bm{\ell}_{q,i}),
\]
where the function \(\texttt{CholeskyFactor}\) transforms the unconstrained vector \(\bm{\ell}_{q,i}\) into a lower-triangular matrix with positive diagonal entries (achieved by exponentiating the diagonal elements), thus ensuring the positive definiteness of \(\bm{\Sigma}_{q,i}\).

To enable gradient-based optimization of the variational parameters \( \bm{\phi} \) in \Cref{eq:iwelbo_improved}, we employ the reparameterization trick to sample $q_{\bm{\phi}}$, that is,
\[
\bm{x}_{i,\text{miss}} = \bm{\mu}_{q,i} + \bm{L}_{q,i} \, \bm{\varepsilon}_i, \qquad 
\bm{\varepsilon}_i \sim \mathcal{N}(\bm{0}, \bm{I}_{d_i}),
\]
where \( \bm{\varepsilon}_i \) is an auxiliary noise variable. 

By amortizing the inference across all observations, the encoder network aims to learn accurate variational parameters for any pattern of observed data \( (\bm{x}_{i,\text{obs}}, y_i) \). This approach significantly reduces the computational burden compared to traditional variational methods that require separate optimization for each data point, while maintaining the flexibility to capture complex dependencies between missing and observed variables.

\section{Extension to MNAR: Modeling Missing Not At Random Mechanisms}
\label{SectionMNAR}
The AV-LR framework originally assumes that the missingness mechanism is ignorable, typically under the \textit{Missing At Random (MAR)} assumption. However, in real-world datasets, this assumption often fails: the probability that a covariate is missing may depend on its unobserved value itself. The mechanism is then non-ignorable and called \textit{Missing Not At Random (MNAR)}.
Figure \ref{fig:graphical_MNAR} depicts the causal relationships among the variables $x$, $y$, and the mask $r$ using a structural causal graph, under a non-ignorable missingness mechanism.

\begin{figure}[!htbp]
\centering
    \begin{tikzpicture}
        \node[obs] (y) {$y$} ; %
        \node[latent, above=of y, path picture={\fill[gray!25] (path picture bounding box.south) rectangle (path picture bounding box.north west);}] (x) {$x$} ; %
        \node[obs, above=of x] (r) {$r$} ; %
        \node[const, right=of x] (musig) {$\mu,\Sigma$} ;
        \node[const, right=of r](psi) {$\psi$};
        \node[const, right=of y](beta) {$\beta$};
        \node[const, left=of x](phi) {$\phi$};
        
        \edge {x} {y} ; %
        \edge {x} {r} ; %
        \path (y) edge[bend left, ->] (r) ;
        \edge {musig}{x};
        \edge {psi}{r};
        \edge {beta}{y};
        \edge[dashed] {phi}{x};
        
        \path (y) edge[dashed, bend right, ->] (x);
        
        \plate {yx} {(y)(x)(r)} {$N$};
    \end{tikzpicture}
         \caption{ Structural causal graph of AV-LR for MNAR covariates. Nodes in grey represent fully observed variables, nodes in white denote unobserved variables, and mixed nodes indicate the presence of both cases. he edges from $x$ to $y$ means that $x$ causes $y$.} 
         \label{fig:graphical_MNAR}
\end{figure}



To capture non‐ignorable missingness (MNAR), we augment our generative model with an explicit stochastic missingness mechanism. The joint likelihood of the observed data and the missingness indicators factorizes as:

\begin{equation}
p_{\bm{\theta}, \bm{\psi}}(y_i, \bm{x}_i, \bm{r}_i) \;=\; 
p_{\bm{\theta}}(y_i \mid \bm{x}_i) \; 
p_{\bm{\theta}}(\bm{x}_i) \; 
p_{\bm{\psi}}(\mathbf{r}_i \mid \bm{x}_i, y_i),
\label{eq:joint_mnar}
\end{equation}

This factorization, called selection models \cite{little2019statistical}, is particularly appropriate when estimating the parameters of the data distribution $(\beta, \mu, \Sigma)$, as it preserves the form of the complete-data distribution while incorporating the missingness mechanism, i.e., the conditional probability of the missingness indicator given the data $p_\psi(\bm{r}_i \mid \bm{x}_i, y_i)$.
The missingness mechanism for each variable is characterized by a Bernoulli distribution, assuming conditional independence of the missingness indicators given $(\bm{x}_i,y_i)$:

\begin{equation}
p_{\boldsymbol{\psi}}(\mathbf{r}_i \mid \mathbf{x}_i, y_i) = 
\prod_{j=1}^d \mathrm{Bernoulli}(r_{ij} \mid \pi_{ij}),
\label{eq:bernoulli_miss}
\end{equation}

where each \( \pi_{ij} \in (0,1) \) is given by:

\begin{equation}
\pi_{ij} = \sigma\big(g_{\bm{\psi}, j}(\bm{x}_i, y_i)\big),
\label{eq:pi_miss}
\end{equation}
with \( g_{\bm{\psi}, j} \) a neural network or linear model parameterized by \( \bm{\psi} \) \cite{lim2024deeply}.
\newcommand{\mask}{\mathbf{m}}

The joint likelihood of the observed quantities becomes
\[
p_{\bm{\theta},\bm{\psi}}(y_i, \bm{x}_{i,\mathrm{obs}}, \bm{r}_i)
= \int p_{\bm{\theta}}(y_i \mid \bm{x}_i)\;p_{\bm{\theta}}(\bm{x}_i)\;p_{\bm{\psi}}(\bm{r}_i \mid \bm{x}_i, y_i)\;\mathrm{d}\bm{x}_{i,\mathrm{miss}}.
\]

Similar to \Cref{eq:ELBO}, by amortizing inference of $\bm{x}_{i,\mathrm{miss}}$ via $q_{\boldsymbol{\phi}}$, the MNAR ELBO for observation $i$ is given by
\begin{align}
\log p_{\boldsymbol{\theta},\boldsymbol{\psi}}\big(y_i,\bm{x}_{i,\mathrm{obs}},\bm{r}_i\big)
&\ge 
\mathbb{E}_{q_{\boldsymbol{\phi}}\!\big(\bm{x}_{i,\mathrm{miss}}\mid \bm{x}_{i,\mathrm{obs}},y_i,\bm{r}_i\big)}
\Big[
\log p_{\boldsymbol{\theta}}\big(y_i\mid \bm{x}_i\big)
+ \log p_{\boldsymbol{\theta}}\big(\bm{x}_i\big)
+ \log p_{\boldsymbol{\psi}}\big(\bm{r}_i\mid \bm{x}_i,y_i\big)\nonumber\\
&\qquad\qquad\qquad\qquad\qquad
- \log q_{\boldsymbol{\phi}}\!\big(\bm{x}_{i,\mathrm{miss}}\mid \bm{x}_{i,\mathrm{obs}},y_i,\bm{r}_i\big)
\Big] \nonumber\\
&=: \; \mathcal{L}_i(\boldsymbol{\theta},\boldsymbol{\psi},\boldsymbol{\phi}).
\label{eq:elbo_mnar}
\end{align}

In practice, as in \Cref{eq:iwelbo_improved}, for each sample \(x_{i,\mathrm{miss}}^{(k)}\sim q_{\bm{\phi}}\), we compute the importance weight
\[
\log w_{i}^{(k)}
= \log p_{\bm{\theta}}(y_i\mid \bm{x}_i^{(k)})
+ \log p_{\bm{\theta}}(\bm{x}_i^{(k)})
+ \log p_{\bm{\psi}}(\bm{r}_i\mid \bm{x}_i^{(k)},y_i)
- \log q_{\bm{\phi}}(\bm{x}_{i,\mathrm{miss}}^{(k)}\mid \bm{x}_{i,\mathrm{obs}},y_i),
\]
and employ the following objective function for training
\begin{equation}
\label{eq:elbo_Mnar}
\mathcal{L}_{\mathrm{IW}}(\bm{\theta}, \bm{\phi}, \bm{\psi})
= \sum_{i=1}^N \log\!\left(\frac{1}{K}\sum_{k=1}^K w_i^{(k)}\right),
\end{equation}
with $w_i^{(k)}=\frac{p_{\bm{\psi}}(\bm{r}_i\mid \bm{x}_i^{(k)},y_i)p_{\bm{\theta}}(y_i\mid \bm{x}_i^{(k)})p_{\bm{\theta}}(\bm{x}_i^{(k)})}{q_{\bm{\phi}}(\bm{x}_{i,\mathrm{miss}}^{(k)}\mid \bm{x}_{i,\mathrm{obs}},y_i)}$.

\section{Prediction with Missing Covariates}
\label{sec:prediction}

The primary objective of this section is to develop a robust prediction framework that properly accounts for missing data mechanisms, enabling accurate probabilistic predictions even when faced with incomplete covariate information. Once the variational encoder \(q_{\bm{\phi}}\) of either form 
\[
q_{\bm{\phi}}(\bm{x}_{i,\mathrm{miss}}\mid \bm{x}_{i,\mathrm{obs}},y_i)\quad(\text{ignorable})
\quad\text{or}\quad
q_{\bm{\phi}}(\bm{x}_{i,\mathrm{miss}}\mid \bm{x}_{i,\mathrm{obs}},y_i,\bm{r}_i)\quad(\text{non‐ignorable}),
\]
has been trained, the target probability is given by:
\begin{equation}\label{eq:corrected_prediction_pi}
\begin{split}
    p_i &= \mathbb{P}(y_i=1\mid\bm{x}_{i,\mathrm{obs}},\bm{r}_i) = \frac{p_{\bm{\theta},\bm{\psi}}(y_i=1,\bm{r}_i\mid\bm{x}_{i,\mathrm{obs}})}{p_{\bm{\theta},\bm{\psi}}(\mathbf{r}_i\mid\bm{x}_{i,\mathrm{obs}})} \\
    &= \frac{\int p_{\bm{\theta}}(y_i=1\mid \bm{x}_{i,\mathrm{obs}},\bm{x}_{i,\mathrm{miss}} ) p_{\bm{\psi}}(\bm{r}_i\mid \bm{x}_{i,\mathrm{obs}},\bm{x}_{i,\mathrm{miss}},y_i=1)p_{\bm{\theta}}(\bm{x}_{i,\mathrm{miss}}\mid \bm{x}_{i,\mathrm{obs}}) \mathrm{d} \bm{x}_{i,\mathrm{miss}} }{
    \sum_{c=0}^1\int p_{\bm{\theta}}(y_i=c\mid \bm{x}_{i,\mathrm{obs}},\bm{x}_{i,\mathrm{miss}} ) p_{\bm{\psi}}(\bm{r}_i\mid \bm{x}_{i,\mathrm{obs}},\bm{x}_{i,\mathrm{miss}},y_i=c)p_{\bm{\theta}}(\bm{x}_{i,\mathrm{miss}}\mid \bm{x}_{i,\mathrm{obs}}) \mathrm{d} \bm{x}_{i,\mathrm{miss}} 
    }.
\end{split}
\end{equation}

In the binary setting $y \in \{0,1\}$, we approximate these integrals via importance sampling using the variational distribution $q_{\phi}$ as the proposal distribution with $S$ draws per class. We now focus on the numerator corresponding to class $1$ (the positive class):

\[
\text{Numerator} \approx \frac{1}{S}\sum_{s=1}^S p_{\mathbf{\theta}}(y_i=1\mid \bm{x}_{i,\mathrm{obs}},\bm{x}_{i,\mathrm{miss}}^{(s,1)} ) p_{\mathbf{\psi}}(\bm{r}_i\mid \bm{x}_{i,\mathrm{obs}},\bm{x}_{i,\mathrm{miss}}^{(s,1)},y_i=1) \frac{p_{\mathbf{\theta}}(\bm{x}_{i,\mathrm{miss}}^{(s,1)}\mid \bm{x}_{i,\mathrm{obs}})}{q_{\mathbf{\phi}}(\bm{x}_{i,\mathrm{miss}}^{(s,1)}\mid \bm{x}_{i,\mathrm{obs}},y_i=1,\bm{r}_i)}
\]
with \(\bm{x}_{i,\mathrm{miss}}^{(s,1)} \sim q_{\mathbf{\phi}}(\bm{x}_{i,\mathrm{miss}}\mid \bm{x}_{i,\mathrm{obs}},y_i=1,\mathbf{r}_i)\).
For the denominator terms (\(c=0,1\)):
\[
\text{Term}_c \approx \frac{1}{S}\sum_{s=1}^S p_{\mathbf{\theta}}(y_i=c\mid \bm{x}_{i,\mathrm{obs}},\bm{x}_{i,\mathrm{miss}}^{(s,c)} ) p_{\mathbf{\psi}}(\bm{r}_i\mid \bm{x}_{i,\mathrm{obs}},\bm{x}_{i,\mathrm{miss}}^{(s,c)},y_i=c) \frac{p_{\mathbf{\theta}}(\bm{x}_{i,\mathrm{miss}}^{(s,c)}\mid \bm{x}_{i,\mathrm{obs}})}{q_{\mathbf{\phi}}(\bm{x}_{i,\mathrm{miss}}^{(s,c)}\mid \bm{x}_{i,\mathrm{obs}},y_i=c,\bm{r}_i)}
\]
where \(\bm{x}_{i,\mathrm{miss}}^{(s,c)} \sim q_{\mathbf{\phi}}(\bm{x}_{i,\mathrm{miss}}\mid \bm{x}_{i,\mathrm{obs}},y_i=c, \bm{r}) \).

The final prediction is computed as:
\[
\hat{p}_i = \frac{\text{Numerator}}{\text{Term}_0 + \text{Term}_1}
\]

\section{Comparison with VAE-based methods}
\label{sec:comparison_VAE}
The distinction between VAE-based approaches for handling missing data and our AV-LR framework lies primarily in their modeling objectives and treatment of incomplete inputs. In standard VAE architectures, the latent variable $\bm{z}$ captures the underlying generative factors of the complete data $\bm{x}$. VAE variants designed for missing data (e.g., MIWAE \cite{mattei2019miwae}, VAEAC \cite{ivanov2018variational}, HI-VAE \cite{nazabal2020handling}) operate as latent-variable generative models that maximize an importance-weighted lower bound:
\begin{equation}
\hat{L}^{\text{vae}}_K
= \sum_{i=1}^n \mathbb{E}_{q_{\phi}(\bm{z}_i,\bm{x}_{i,\text{mis}}\mid \bm{x}_{i,\text{obs}},\bm{r}_i)}\!\left[
\log\!\left(\frac{1}{K}\sum_{k=1}^K w_{i k}\right)\right],
\end{equation}
\[
w_{i k} \;=\;
\frac{p_{\psi}\big(\bm{r}_i \mid \bm{x}_{i,\text{obs}},\, \bm{x}_{i,\text{mis}}^{(k)}\big)\;
      p_{\theta}\big(\bm{x}_{i,\text{obs}}\mid \bm{z}_i^{(k)}\big)\;
      p\big(\bm{z}_i^{(k)}\big)}
     {q_{\phi}\big(\bm{z}_i^{(k)}\mid \bm{x}_{i,\text{obs}},\bm{r}_i\big)},
\qquad k=1,\dots,K,
\]
 These methods prioritize learning flexible posteriors $q_{\phi}(\bm{z}\mid\bm{x}_{\text{obs}},\bm{r})$ to reconstruct $\bm{x}_{\text{mis}}$ (while optionally modeling $p_{\psi}(\bm{r}\mid\cdot)$ under MNAR), making them fundamentally optimized for data reconstruction rather than downstream predictive tasks.

Recent hybrid approaches like DLGLM (Deep Latent Gaussian Linear Model) \cite{lim2024deeply} attempt to bridge this gap by integrating VAEs with generalized linear models for supervised prediction. DLGLM employs a deep neural network to model the relationship between covariates and response, i.e., $p(y|\bm{x}) = \text{GLM}(y; \eta(\bm{x}))$ where $\eta(\bm{x}) = s_{\beta,\pi}(\bm{x})$ is a neural network. Simultaneously, it models the covariate distribution $p(\bm{x})$ using an IWAE with latent variables $\bm{z}$. In the presence of missing covariates, DLGLM introduces a variational distribution $q_{\theta}(\bm{z}, \bm{x}_{\text{mis}} | \bm{x}_{\text{obs}}, \bm{r})$ that factorizes as $q_{\theta_1}(\bm{z} | \bm{x}_{\text{obs}}) q_{\theta_2}(\bm{x}_{\text{mis}} | \bm{z}, \bm{x}_{\text{obs}}, \bm{r})$. The model is trained end-to-end by maximizing a lower bound (the dlglm bound) that involves importance-weighted sampling over both $\bm{z}$ and $\bm{x}_{\text{mis}}$:
\[
\hat{L}^{\text{dlglm}}_K = \sum_{i=1}^n \log \frac{1}{K} \sum_{k=1}^K \frac{p_{\alpha,\beta,\pi}(y_i | \bm{x}_{i,\text{obs}}, \tilde{\bm{x}}_{i,\text{mis}}^{(k)}) \, p_{\psi}(\bm{x}_{i,\text{obs}}, \tilde{\bm{x}}_{i,\text{mis}}^{(k)} | \tilde{\bm{z}}_i^{(k)}) \, p(\tilde{\bm{z}}_i^{(k)}) \, p_{\phi}(\bm{r}_i | \bm{x}_{i,\text{obs}}, \tilde{\bm{x}}_{i,\text{mis}}^{(k)}, y_i)}{q_{\theta_1}(\tilde{\bm{z}}_i^{(k)} | \bm{x}_{i,\text{obs}}) \, q_{\theta_2}(\tilde{\bm{x}}_{i,\text{mis}}^{(k)} | \tilde{\bm{z}}_i^{(k)}, \bm{x}_{i,\text{obs}}, \bm{r}_i)},
\]
where samples are drawn as $\tilde{\bm{z}}_i^{(k)} \sim q_{\theta_1}(\bm{z}_i | \bm{x}_{i,\text{obs}},y_i)$ and $\tilde{\bm{x}}_{i,\text{mis}}^{(k)} \sim q_{\theta_2}(\bm{x}_{i,\text{mis}} | \tilde{\bm{z}}_i^{(k)}, \bm{x}_{i,\text{obs}}, \bm{r}_i, y_i)$. This results in a joint learning framework that simultaneously handles missing data imputation and prediction.

However, this approach requires Monte Carlo approximations over both the latent variables and the missing covariates. While Monte Carlo convergence rates are dimension-independent in theory, the nested sampling scheme (first $\bm{z}$, then $\bm{x}_{\text{mis}}$ conditioned on $\bm{z}$) introduces additional computational complexity and may require careful tuning of the variational distributions to maintain efficiency.

In contrast, AV-LR directly targets the joint likelihood
\[
p_{\theta}(y, \bm{x}_{\text{obs}}, \bm{x}_{\text{mis}}) \quad (\text{extended to } p_{\theta,\psi}(y, \bm{x}, \bm{r}) \text{ under MNAR}),
\]
through a streamlined variational framework that approximates the posterior over missing covariates directly:
\[
q_{\phi}(\bm{x}_{\text{mis}} \mid \bm{x}_{\text{obs}}, y).
\]
Our evidence lower bound decomposes as:
\[
\mathbb{E}_{q_{\phi}}[\log p_{\theta}(y \mid \bm{x}_{\text{obs}}, \bm{x}_{\text{mis}})] + \mathbb{E}_{q_{\phi}}[\log p_{\theta}(\bm{x}_{\text{obs}}, \bm{x}_{\text{mis}})] - \mathbb{E}_{q_{\phi}}[\log q_{\phi}(\bm{x}_{\text{mis}} \mid \bm{x}_{\text{obs}}, y)],
\]
enabling truly joint learning of imputation and classification parameters \cite{kingma2013auto}. This formulation yields critical advantages over both DLGLM and standard VAEs. By eliminating intermediate latent variables and directly modeling $q_{\phi}(\bm{x}_{\text{mis}} \mid \bm{x}_{\text{obs}}, y)$, AV-LR: (1) avoids the need for joint sampling over $\bm{z}$ and $\bm{x}_{\text{mis}}$, (2) decouples imputation from latent representation learning, (3) accommodates flexible covariate models without complicating the predictive pathway, and (4) simplifies the optimization by reducing the number of stochastic layers in the computational graph.

\section{Simulation Study}
\label{SectionExperiments}

\paragraph{Measuring the performances} We evaluate the performance of all methods using multiple criteria that assess different aspects of model quality. For parameter estimation accuracy (see \ref{LR_model_binary} and \ref{log_obs_binary} for model setup), we compute the root mean square error (RMSE) between estimated and true values for: (i) the regression coefficients $\boldsymbol{\beta}$, (ii) the covariate mean vector $\boldsymbol{\mu}$, and (iii) the imputed missing values. For covariance matrix recovery, we use the Frobenius norm of the discrepancy between the estimated and true covariance matrices. Predictive performance is assessed through classification metrics including area under the ROC curve (AUC-ROC), accuracy, precision, recall, and F1 score. For probabilistic calibration, we employ the Brier score. Finally, computational efficiency is measured by training and testing times. All metrics are computed over multiple random seeds to ensure statistical reliability, with results reported as mean   standard deviation.

\subsection{The gain to take into account the non-ignorable nature of the mechanism}
\label{subsec:behaviour_avlr}

\begin{figure}[htbp]
  \centering
  \includegraphics[width=\linewidth]{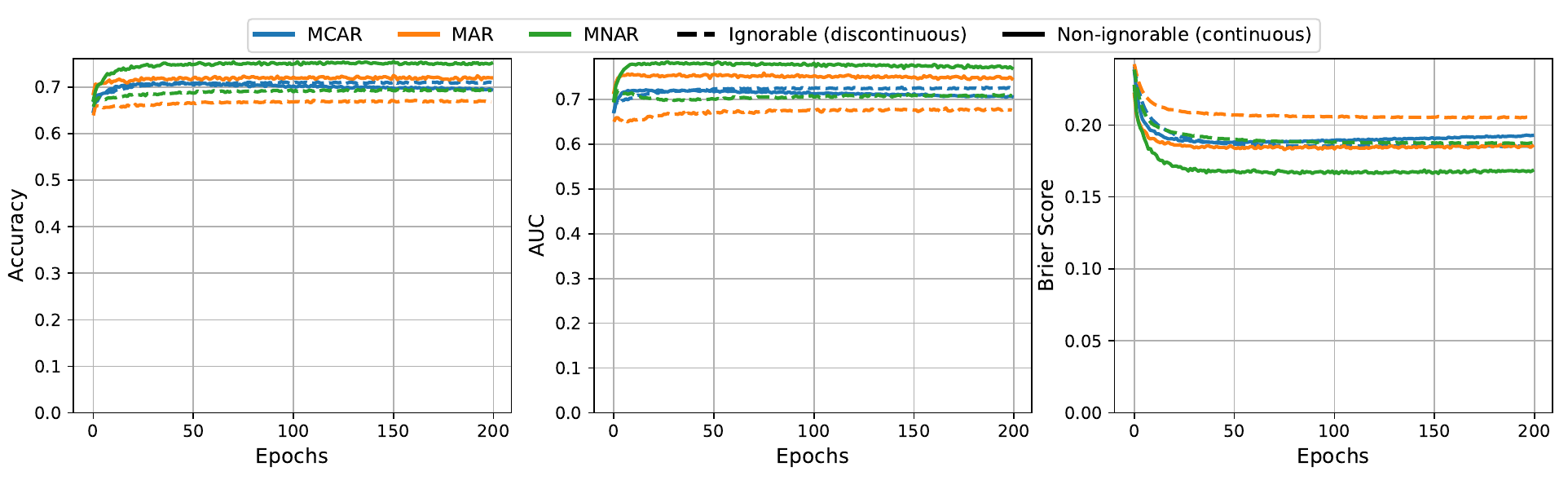}
  \caption{Evolution of classification metrics 
    (accuracy, AUC, and Brier score) over epochs. 
    The solid line corresponds to AV‑LR (non‑ignorable extension) and the dashed line to AV‑LR (ignorable). 
    Colors denote missingness mechanisms: blue for 50\% MCAR, orange for 60\% MAR, and green for 60\%\ MNAR. 
    Synthetic data were generated with $(n,p)=(5000,5)$ for training and $(1000,5)$ for testing.}
  \label{fig:classification_metrics}
\end{figure}

This first experiment evaluates the robustness of AV-LR across three missingness mechanisms. 
The assessment is conducted on synthetic datasets with $(N,d) = (5000,5)$ for training 
and $(1000,5)$ for testing, over 200 epochs and averaged across five runs:

\begin{itemize}
    \item \textbf{MCAR (50\%)}: Each entry $x_{ij}$ is independently missing with 
    probability $p=0.5$, i.e., $r_{ij}\sim\text{Bernoulli}(0.5)$.
    
 \item \textbf{MAR (60\%)}: Missingness in all variables depends on the first covariate $x_{i,1}$ and the response variable $y_i$ via a logistic model:
$$p(r_{ij}=1\mid \bm{x}_i, y_i) = 
    \sigma(-1.0 + 1.5 \cdot x_{i,1} - 0.8 \cdot y_i),$$
where \(\sigma(z) = \frac{1}{1+e^{-z}}\) is the logistic function.
The same missingness pattern is applied to all features, with coefficients calibrated to achieve approximately 60\% of missingness per variable.

\item \textbf{MNAR (60\%)}: Missingness depends on all covariate values and the response variable:
\[
p(r_{ij}=1\mid \boldsymbol{x}_i,y_i) = 
\sigma\left(-1.0 -2.0 \cdot \sum_{k=1}^{d} x_{ik} + 0.5 \cdot y_i\right),
\]
where the missingness of each feature depends on the values of all covariates in the observation and the response, with parameters calibrated to yield approximately 60\% of missingness per variable.
\end{itemize}

\paragraph{Classification performance}
Under the \textsc{MNAR} mechanism, the non-ignorable extension of AV-LR exhibits consistent and substantial performance improvements, observed in AUC, accuracy, and the Brier score, relative to the ignorable variant (cf. Fig.~\ref{fig:classification_metrics}). These gains stem primarily from improved calibration of predicted probabilities and finer discrimination between classes, resulting from an effective correction of the sampling bias induced by non-random missingness. The trajectories over the 200 training epochs indicate that the performance gap appears early in training and persists through convergence, suggesting that explicit modeling of the non-response mechanism yields a sustained improvement in estimating $p(\bm{y}\mid \bm{X})$ in this setting.

\paragraph{Mechanism-Dependent Learning Dynamics}
The superiority of the non-ignorable extension under \textcolor{green}{MNAR} stems from its explicit modeling of the missingness mechanism $p(\bm{r}|\bm{x},\bm{y})$. Surprisingly, it also shows advantages under \textcolor{orange}{MAR} despite the theoretically ignorable nature, suggesting practical benefits from modeling covariate-missingness correlations. Under \textcolor{blue}{MCAR}, performance parity confirms that both variants are equally suitable for random missingness.

\subsection{Comparison with existing methods}


\noindent In this simulation study, we systematically evaluate the performance of various imputation and direct estimation methods for binary logistic regression in the presence of missing covariate data. Synthetic datasets are generated according to a fully reproducible protocol: a training set of $N_{\mathrm{train}} = 2000$ observations and a test set of $N_{\mathrm{test}} = 500$ observations, each comprising $d = 5$ covariates drawn from a multivariate normal distribution $\mathcal{N}(\boldsymbol{\mu}, \boldsymbol{\Sigma})$. 
The true regression coefficients $\boldsymbol{\beta}$, mean vector $\boldsymbol{\mu}$, and covariance matrix $\boldsymbol{\Sigma}$ are held fixed a priori and serve as ground‐truth benchmarks. Estimation accuracy is assessed by comparing the methods’ parameter estimates to these benchmarks, and predictive performance is evaluated on the independent test set.

\paragraph{Benchmark and competitive methods} Monte Carlo simulations are conducted to ensure the stability and robustness of our findings by comparing seven estimators: simple mean imputation, MICE (Multiple Imputation by Chained Equations) \cite{van2011mice}, missForest (random forest imputation) \cite{stekhoven2012missforest}, MIWAE (Missing data Importance-Weighted Autoencoder)\cite{mattei2019miwae}, DLGLM (Deeply Learned Generalized Linear Model) for MNAR or IDLGLM (Ignorable DLGLM) for MAR/MCAR mechanisms \cite{lim2024deeply}, the Stochastic Approximation Expectation–Maximization (SAEM) algorithm \cite{jiang2020logistic}, and our proposed AV-LR method, which embeds amortized variational inference directly within the logistic regression framework.  For comparability, MICE is implemented as a single-imputation procedure, consistent with the other methods considered. For the first four methods (mean, MICE, missForest, MIWAE), missing values in the test set are imputed independently from the training phase, and the estimated coefficients \(\boldsymbol{\beta}\) obtained during training are then applied to the imputed test covariates for prediction; conversely, SAEM, DLGLM/IDLGLM, and AV-LR natively handle missing data at prediction time without requiring separate preprocessing of the test set. 

\paragraph{Hyperparameters tuning} For the neural network-based methods (MIWAE, IDLGLM, and AV-LR), we use a consistent architecture: a single hidden layer with 128 units. These models are trained for 150 epochs with a batch size of 256 and a learning rate of $10^{-3}$. The SAEM algorithm is run for a maximum of 120 iterations, with a convergence threshold set to $10^{-4}$. For IDLGLM, we adopt the architecture and training configuration described above, while other hyperparameters are set to the values recommended as optimal in the original study~\cite{lim2024deeply}.

\begin{table}[htbp]
\centering
\caption{Average and standard deviation of parameter RMSE, AUC and Accuracy (5 repetitions) under 50\% MCAR mechanism}
\label{tab:param_rmse_50mcar}
\arrayrulecolor{blue}
\renewcommand{\arraystretch}{1.15}
\rowcolors{2}{blue!8}{blue!3} 
\resizebox{\textwidth}{!}{%
\begin{tabular}{lrrrr|rr}
\rowcolor{blue!30}
\textcolor{white}{Estimator} &
\textcolor{white}{RMSE\_imp} &
\textcolor{white}{RMSE\_mu} &
\textcolor{white}{Frobenius\_cov} &
\textcolor{white}{RMSE\_beta} &
\textcolor{white}{AUC} &
\textcolor{white}{Accuracy} \\
\hline
AV-LR       & \textbf{0.8029 $\pm$ 0.0181} & \textbf{0.0081 $\pm$ 0.0023} & \textbf{0.1280 $\pm$ 0.0312} & \textbf{0.0854 $\pm$ 0.0035} & \textbf{0.7420 $\pm$ 0.0850} & \textbf{0.7375 $\pm$ 0.0490} \\
IDLGLM      & 0.8325 $\pm$ 0.0214 & 0.0108 $\pm$ 0.0021 & 0.3527 $\pm$ 0.0183 & 0.2471 $\pm$ 0.0416 & 0.7380 $\pm$ 0.0720 & 0.7290 $\pm$ 0.0560 \\
MIWAE       & 0.8166 $\pm$ 0.0190 & 0.0135 $\pm$ 0.0027 & 0.4158 $\pm$ 0.0142 & 0.2983 $\pm$ 0.0498 & 0.7280 $\pm$ 0.0700 & 0.7200 $\pm$ 0.0520 \\
SAEM        & 0.8055 $\pm$ 0.0187 & 0.0129 $\pm$ 0.0030 & 0.3650 $\pm$ 0.0125 & 0.2714 $\pm$ 0.0720 & 0.7305 $\pm$ 0.0802 & 0.7189 $\pm$ 0.0601 \\
MICE        & 0.8699 $\pm$ 0.0175 & 0.0149 $\pm$ 0.0042 & 0.4820 $\pm$ 0.0103 & 0.4786 $\pm$ 0.0801 & 0.7250 $\pm$ 0.0888 & 0.7123 $\pm$ 0.0705 \\
MissForest  & 0.8859 $\pm$ 0.0163 & 0.0205 $\pm$ 0.0050 & 0.3921 $\pm$ 0.0154 & 0.5052 $\pm$ 0.0187 & 0.7190 $\pm$ 0.0950 & 0.7155 $\pm$ 0.0620 \\
Mean        & 1.0183 $\pm$ 0.0210 & 0.0158 $\pm$ 0.0041 & 2.3051 $\pm$ 0.0375 & 0.4521 $\pm$ 0.0612 & 0.7120 $\pm$ 0.0920 & 0.7012 $\pm$ 0.0750 \\
\end{tabular}%
}
\end{table}

\begin{table}[htbp]
\centering
\caption{Average and standard deviation of execution times under 50\% MCAR (5 repetitions)}
\label{tab:exec_time_50mcar}
\begin{tabular}{lrr}
\toprule
Estimator   & Training Time    & Testing Time     \\
\midrule
AV-LR    & 34.4793 $\pm$ 4.2157 & 0.0156 $\pm$ 0.0042   \\
IDLGLM       & 63.5544 $\pm$ 9.0132 & 8.0829 $\pm$ 1.4923   \\
MIWAE       & 35.4845 $\pm$ 4.9321 & 2.0263 $\pm$ 0.3187   \\
SAEM        & 3894.93$\pm$ 166.4353 & 34.1459 $\pm$ 6.1487   \\
MICE        & 0.1852 $\pm$ 0.0320 & 0.1627 $\pm$ 0.0410   \\
missForest  & 38.3445 $\pm$ 6.1023 & 24.4828 $\pm$ 3.1124  \\
Mean        & 0.0429 $\pm$ 0.0051 & 0.0010 $\pm$ 0.0002   \\
\bottomrule
\end{tabular}
\end{table}

\subsubsection{Ignorable Missing Data}

We conducted a comprehensive evaluation of synthetic datasets under MCAR with a 50\% missingness level. AV-LR consistently outperformed the baseline methods in both parameter recovery and predictive accuracy (see Tables~\ref{tab:param_rmse_50mcar}--\ref{tab:exec_time_50mcar}). At 50\% missingness, AV-LR maintained superior parameter fidelity, with an imputation RMSE of approximately 0.80, and achieved the best or near-best predictive performance across the evaluated metrics. In terms of computation, AV-LR required a moderate training time in our experiments (on the order of a few tens of seconds) while offering very low inference latency (around $10^{-2}$ seconds). In contrast, SAEM exhibited a substantially higher training cost (on the order of $10^{3}$ seconds), highlighting the significant practical speed advantage of AV-LR. Compared to MICE, AV-LR incurs a higher computational cost during training due to its greater model complexity, which is required to fit stochastic regressions. In contrast, the testing phase is more efficient for AV-LR, since predictions are obtained directly without requiring additional imputation steps.

\subsubsection{Non‑Ignorable Missing Data}

\noindent In this second set of experiments, we evaluate the same suite of methods under an MNAR  mechanism in which missingness depends both on the covariate values and the binary outcome.  Synthetic datasets are generated as before with \(N_{\mathrm{train}}=2000\), \(N_{\mathrm{test}}=500\) and \(p=5\) covariates drawn from \(\mathcal{N}(\boldsymbol{\mu},\boldsymbol{\Sigma})\).  For simplicity, we adopt this logistic missingness model and assume that the indicators \(r_{ij}\) are conditionally independent given the full covariate vector \(\bm{x}_i\) and outcome \(y_i\):
\[
\Pr(r_{ij}=1 \mid \bm{x}_{i},y_i)
=\sigma\bigl(\psi_{0j} + \bm{\psi}_{1j}^\top \bm{x}_{i} + \psi_{2j}\,y_i\bigr),
\]
where the feature-specific parameters \(\{\psi_{0j},\bm{\psi}_{1j},\psi_{2j}\}\) are calibrated to yield approximately 60\% overall missingness.  

\noindent The baseline methods (mean, MICE \cite{van2011mice}, missForest \cite{stekhoven2012missforest}, SAEM \cite{jiang2020logistic}) are identical to those used in the ignorable case. In addition, three extensions specifically adapted to the MNAR mechanism are considered: Not‑MIWAE \cite{ipsen2020not}, DLGLM, and AV‑LR MNAR. Not‑MIWAE \cite{ipsen2020not} is an extension of MIWAE \cite{mattei2019miwae} that explicitly models the missingness mechanism \(p(r\mid x,y)\). DLGLM is the non‑ignorable variant of IDLGLM \cite{lim2024deeply} adjusted to account for dependence of \(r\) on both the covariates and the response, and AV-LR MNAR is our proposed amortized variational inference method extended to handle MNAR (see Section \eqref{SectionMNAR}).

Under the demanding 60\% MNAR setting, AV-LR consistently outperforms competing estimators in both parameter recovery and predictive accuracy (Tables~\ref{tab:param_rmse_60mnar}--\ref{tab:exec_time_60mnar}). In terms of reconstruction, AV-LR achieves markedly lower errors for imputation, mean, and covariance estimation, whereas classical methods such as SAEM and MICE exhibit substantially higher deviations. 

On the predictive side, AV-LR achieves the highest overall AUC and accuracy scores, indicating superior performance. Although missForest shows competitive performance on some metrics, its performance is less consistent across all indicators. All metrics used for this evaluation are provided in the Appendix \eqref{app:1}.

In addition, AV-LR requires only moderate training and very low inference time, in contrast to the prohibitive computational cost of SAEM. Overall, these results underline AV-LR as the most robust and efficient approach under severe non-ignorable missingness.

\begin{table}[htbp]
\centering
\caption{Average and standard deviation of parameter RMSE, AUC and Accuracy (5 repetitions) under 60\% MNAR}
\label{tab:param_rmse_60mnar}
\arrayrulecolor{blue}
\renewcommand{\arraystretch}{1.12}
\rowcolors{2}{blue!8}{blue!3} 
\resizebox{\textwidth}{!}{%
\begin{tabular}{lrrrr|rr}
\rowcolor{blue!30}
\textcolor{white}{Estimator} &
\textcolor{white}{RMSE\_imp} &
\textcolor{white}{RMSE\_mu} &
\textcolor{white}{Frobenius\_cov} &
\textcolor{white}{RMSE\_beta} &
\textcolor{white}{AUC} &
\textcolor{white}{Accuracy} \\
\hline
AV-LR       & \textbf{0.8624 $\pm$ 0.0300} & \textbf{0.3300 $\pm$ 0.0200} & \textbf{1.3000 $\pm$ 0.0800} & \textbf{0.5400 $\pm$ 0.2100} & \textbf{0.771 $\pm$ 0.060} & \textbf{0.724 $\pm$ 0.050} \\
DLGLM       & 0.8800 $\pm$ 0.0350 & 0.3400 $\pm$ 0.0200 & 1.3500 $\pm$ 0.0850 & 0.5600 $\pm$ 0.2200 & 0.768 $\pm$ 0.055 & 0.716 $\pm$ 0.045 \\
NotMIWAE    & 0.9200 $\pm$ 0.0300 & 0.3600 $\pm$ 0.0200 & 1.4200 $\pm$ 0.1100 & 0.6000 $\pm$ 0.2300 & 0.738 $\pm$ 0.117 & 0.682 $\pm$ 0.097 \\
SAEM        & 1.3000 $\pm$ 0.0200 & 0.6200 $\pm$ 0.0250 & 2.0000 $\pm$ 0.0800 & 0.6600 $\pm$ 0.2900 & 0.625 $\pm$ 0.063 & 0.600 $\pm$ 0.045 \\
MICE        & 1.2579 $\pm$ 0.0132 & 0.5791 $\pm$ 0.0036 & 2.0530 $\pm$ 0.0849 & 0.5545 $\pm$ 0.3647 & 0.659 $\pm$ 0.045 & 0.520 $\pm$ 0.051 \\
missForest  & 1.0749 $\pm$ 0.0624 & 0.6280 $\pm$ 0.0382 & 2.0712 $\pm$ 0.1309 & 0.7120 $\pm$ 0.3950 & 0.725 $\pm$ 0.058 & 0.666 $\pm$ 0.060 \\
Mean        & 1.4825 $\pm$ 0.0222 & 0.7354 $\pm$ 0.0075 & 2.9295 $\pm$ 0.0188 & 0.7709 $\pm$ 0.3418 & 0.634 $\pm$ 0.078 & 0.508 $\pm$ 0.054 \\
\end{tabular}%
}
\end{table}

\begin{table}[htbp]
\centering
\small
\setlength{\tabcolsep}{6pt}
\caption{Average and standard deviation of execution times under 60\% MNAR}
\label{tab:exec_time_60mnar}
\begin{tabular}{lrr}
\toprule
Estimator   & Training Time (s) & Testing Time (s) \\
\midrule
AV-LR & 62.915 $\pm$ 6.292  & 0.156 $\pm$ 0.016  \\
DLGLM      & 80.556 $\pm$ 8.056  & 6.078 $\pm$ 0.608  \\
NotMIWAE      & 44.607 $\pm$ 4.461  & 1.232 $\pm$ 0.123  \\
SAEM          & 4211.332 $\pm$ 181.877 & 36.433 $\pm$ 6.511 \\
MICE          & 0.118 $\pm$ 0.020   & 0.079 $\pm$ 0.010  \\
missForest    & 30.708 $\pm$ 5.000  & 14.315 $\pm$ 2.500 \\
Mean          & 0.012 $\pm$ 0.002   & 0.000 $\pm$ 0.000  \\
\bottomrule
\end{tabular}
\end{table}

\section{Real-World Dataset Evaluation}
\label{sec:real_world}

\subsection{Several complete real datasets}

We evaluated AV-LR on four diverse benchmark datasets to assess its performance across various domains and missingness scenarios. Each dataset presents distinct characteristics that challenge missing data handling methods:

\begin{itemize}
  \item \textbf{BankNote Authentication}: 1,372 samples, 4 wavelet features (variance, skewness, kurtosis, entropy), high inter-feature correlations ($|\rho| \leq 0.86$).
  
  \item \textbf{Pima Indians Diabetes}: 768 samples, 8 clinical predictors.
  \item \textbf{Rice Cammeo Osmancik}: 3,810 samples, 7 morphological grain measurements, complex nonlinear relationships.
  
  \item \textbf{Breast Cancer Diagnostic}: 569 samples, 30 radiomic features, high dimensionality with limited samples.
\end{itemize}
\subsection*{Missingness Mechanisms}

In our experiments, we simulated three structured MNAR mechanisms to introduce approximately 50\% missingness in the data. These mechanisms follow established definitions in the missing-data literature.

\begin{enumerate}
  \item \textbf{Self-Masking Mechanism} \\
  Here, for each covariate \(j\) and sample \(i\), the probability of missingness depends only on the value of that covariate itself :
  \[
    p(r_{ij} = 1 \mid x_{ij}) = \sigma\!\left( \psi_{j0} + \psi_{j1} \cdot x_{ij} \right).
  \]
   This models scenarios where extreme values are preferentially missing, consistent with the ``self-masked'' MNAR mechanism described in the literature \cite{mohan2018handling}.

  \item \textbf{Logistic Mechanism} \\
  Missingness depends jointly on all covariates and the response:
  \[
    p(r_{ij} = 1 \mid \boldsymbol{x}_i, y_i) = \sigma\!\left( \psi_{j0} + \sum_{k=1}^d \psi_{jk}\,x_{ik} + \psi_{j(d+1)}\,y_i \right)
  \]
  This mechanism reflects systemic biases where missingness relates to both feature values and the outcome variable, similar to MAR-augmented logistic models or MNAR-Y setups  \cite{rubin1987statistical} .

  \item \textbf{Sequential Logistic (Reverse‑Chain) Mechanism} \\
  To capture inter-feature dependencies in missingness, we generate missingness indicators sequentially in reverse order from \(j=d\) down to \(j=1\):
  \[
\begin{aligned}
  \eta_{ij} &= \psi_{j0} + \psi_{j1}\,y_i
    + \sum_{k=1}^{d} \psi_{j,\,1+k}\,x_{ik}
    + \sum_{k=j+1}^{d} \psi_{j,\,1+d+k}\,r_{ik},\ &
  p\bigl(r_{ij}=1 \mid \bm{x}_i, y_i, \bm{r}_{i,j+1:d}\bigr)
    &= \sigma\bigl(\eta_{ij}\bigr)
\end{aligned}
\]

  Thus, the missingness of feature \(j\) depends not only on covariates and the response, but also on the previously drawn missingness indicators of features with indices greater than \(j\). This simulates cascading missingness patterns where the recording (or omission) of one feature affects others, an approach often used to mimic complex MNAR processes \cite{sadinle2018sequential,ibrahim1999missing}.
\end{enumerate}
These three mechanisms span a spectrum of MNAR settings: from self-dependent censoring (Self‑Masking) to outcome-correlated missingness (Logistic), to interdependent cascade patterns (Sequential Logistic). This diversity allows evaluating methods in controlled yet realistic MNAR scenarios consistent with modern missing-data literature.

\noindent Table~\ref{tab:auc_transposed_50mnar} presents AUC scores for every dataset and MNAR mechanism considered. AV-LR consistently attains the highest AUC across datasets and mechanisms. SAEM is not included in the comparison, as it is computationally expensive and not designed to handle MNAR data. The full set of predictive measures (accuracy, precision, specificity, F1-score, Brier score, etc.) for all methods, together with detailed execution-time tables, are provided in the Appendix \eqref{all_REAL}. These extended results corroborate AV-LR's superior and stable predictive performance, while alternative methods occasionally exhibit strengths on specific datasets or metrics.

\begin{table}[htbp]
\centering
\caption{Comparative performance analysis of imputation methods under 50\% MNAR missingness: AUC scores (mean $\pm$ standard deviation)}
\label{tab:auc_transposed_50mnar}
\small
\setlength{\tabcolsep}{4pt}
\begin{tabular}{lcccccc}
\toprule
\rowcolor{gray!20}
Dataset / Mechanism & AV-LR & DLGLM & NotMIWAE & MICE & MissForest & Mean \\
\midrule
\multicolumn{7}{l}{\textit{BankNote}} \\
\quad Self-Masking & \textbf{0.873 $\pm$ 0.024} & 0.853 $\pm$ 0.021 & 0.839 $\pm$ 0.020 & 0.772 $\pm$ 0.034 & 0.745 $\pm$ 0.049 & 0.760 $\pm$ 0.035 \\
\quad Logistic & \textbf{0.804 $\pm$ 0.040} & 0.765 $\pm$ 0.028 & 0.718 $\pm$ 0.027 & 0.587 $\pm$ 0.022 & 0.569 $\pm$ 0.065 & 0.560 $\pm$ 0.040 \\
\quad Sequential Logistic & \textbf{0.842 $\pm$ 0.026} & 0.803 $\pm$ 0.029 & 0.775 $\pm$ 0.028 & 0.802 $\pm$ 0.074 & 0.828 $\pm$ 0.022 & 0.680 $\pm$ 0.040 \\

\addlinespace[2pt]
\multicolumn{7}{l}{\textit{Pima}} \\
\quad Self-Masking & 0.703 $\pm$ 0.048 & 0.741 $\pm$ 0.044 & \textbf{0.748 $\pm$ 0.050} & 0.632 $\pm$ 0.110 & 0.730 $\pm$ 0.059 & 0.568 $\pm$ 0.059 \\
\quad Logistic & \textbf{0.748 $\pm$ 0.043} & 0.725 $\pm$ 0.036 & 0.746 $\pm$ 0.051 & 0.705 $\pm$ 0.046 & 0.698 $\pm$ 0.056 & 0.690 $\pm$ 0.055 \\
\quad Sequential Logistic & \textbf{0.787 $\pm$ 0.028} & 0.748 $\pm$ 0.027 & 0.753 $\pm$ 0.040 & 0.730 $\pm$ 0.044 & 0.757 $\pm$ 0.029 & 0.715 $\pm$ 0.042 \\

\addlinespace[2pt]
\multicolumn{7}{l}{\textit{Rice}} \\
\quad Self-Masking & \textbf{0.953 $\pm$ 0.005} & 0.934 $\pm$ 0.010 & 0.943 $\pm$ 0.005 & 0.951 $\pm$ 0.004 & 0.948 $\pm$ 0.007 & 0.626 $\pm$ 0.026 \\
\quad Logistic & \textbf{0.977 $\pm$ 0.003} & 0.973 $\pm$ 0.005 & 0.954 $\pm$ 0.008 & 0.969 $\pm$ 0.004 & 0.971 $\pm$ 0.003 & 0.945 $\pm$ 0.013 \\
\quad Sequential Logistic & \textbf{0.972 $\pm$ 0.004} & 0.967 $\pm$ 0.004 & 0.959 $\pm$ 0.005 & 0.967 $\pm$ 0.005 & 0.970 $\pm$ 0.003 & 0.940 $\pm$ 0.014 \\

\addlinespace[2pt]
\multicolumn{7}{l}{\textit{BreastCancer}} \\
\quad Self-Masking & 0.981 $\pm$ 0.011 & 0.976 $\pm$ 0.013 & 0.971 $\pm$ 0.012 & \textbf{0.985 $\pm$ 0.006} & 0.980 $\pm$ 0.012 & 0.958 $\pm$ 0.021 \\
\quad Logistic & \textbf{0.986 $\pm$ 0.006} & 0.979 $\pm$ 0.006 & 0.971 $\pm$ 0.005 & 0.977 $\pm$ 0.006 & 0.974 $\pm$ 0.010 & 0.956 $\pm$ 0.008 \\
\quad Sequential Logistic & \textbf{0.982 $\pm$ 0.011} & 0.974 $\pm$ 0.009 & 0.966 $\pm$ 0.008 & 0.967 $\pm$ 0.010 & 0.975 $\pm$ 0.009 & 0.949 $\pm$ 0.016 \\
\bottomrule
\end{tabular}
\end{table}

\subsection{Incomplete dataset: the NHANES dataset}

The National Health and Nutrition Examination Survey (NHANES) is a comprehensive research program conducted by the National Center for Health Statistics (NCHS) to assess the health and nutritional status of the United States population. This program uniquely combines interview-based data with physical examinations, creating a rich multidimensional dataset that has supported thousands of research publications.
For our experimental analysis, we utilized a curated subset of the NHANES 2013--2014 data containing 8,530 complete observations with 13 continuous covariates selected from the various examination components. The selected subset exhibited a missing data rate of 26\% across these covariates, reflecting the authentic patterns of incomplete data in the original NHANES dataset. These covariates encompass key health indicators, including anthropometric measurements, blood pressure readings, and biochemical markers from laboratory tests.
The target variable for our classification task exhibits a natural distribution of approximately 80\% negative cases (class 0) and 20\% positive cases (class 1), reflecting the actual prevalence patterns observed in the population. We preserved this natural class distribution without artificial rebalancing to maintain the authentic characteristics of the dataset.
Our experimental design incorporated two distinct methodological approaches to handle missing data. The first approach operated under the assumption of ignorable missingness, while the second addressed the more complex scenario of non-ignorable missingness. In both experimental frameworks, we utilized the naturally occurring missing data patterns present in the original NHANES dataset without introducing artificial missing values, ensuring that our evaluation reflects realistic conditions with authentic missing data mechanisms across the selected 13 continuous covariates.

\subsection*{Results under the Ignorable Missingness Assumption}

\noindent Under the ignorable missingness assumption, AV-LR provides the best overall trade-off between calibration and discrimination, with stable performance across repetitions (see Table~\ref{tab:ignorable_comparison}). Some competing approaches (notably MICE) can reach comparable accuracy but tend to exhibit reduced calibration or greater variability; SAEM achieves high sensitivity at the expense of substantially higher computational cost. Simple mean imputation performs consistently worse than principled methods, underscoring the importance of dedicated missing-data modeling. Overall, AV-LR emerges as a pragmatic choice that balances predictive quality and computational efficiency.

\subsection*{Results under the Non-Ignorable Missingness Assumption}

\noindent Under the non-ignorable missingness assumption, the principal trends are preserved: AV-LR remains robust and maintains strong discriminative ability (see Table~\ref{tab:nonignorable_comparison}). Alternative methods (e.g., MIWAE or MICE) may outperform on specific metrics in particular configurations but often exhibit greater variability. Naive mean imputation continues to underperform, confirming that structured approaches are preferable when missingness is present. In practice, modelling non-ignorability increases training cost for some methods, while AV-LR retains a favourable performance-to-cost ratio.

\begin{table}[htbp]
\centering
\caption{Comparative performance under ignorable missingness on NHANES dataset}
\label{tab:ignorable_comparison}
\scriptsize
\setlength{\tabcolsep}{3pt}
\begin{tabular}{lcccccc}
\toprule
\textbf{Method} & \textbf{Brier Score} & \textbf{AUC} & \textbf{Accuracy} & \textbf{Precision} & \textbf{Recall} & \textbf{F1-Score} \\
\midrule
AV-LR & \textbf{0.194(0.005)} & \textbf{0.776(0.007)} & 0.694(0.012) & \textbf{0.369(0.010)} & \textbf{0.749(0.019)} & \textbf{0.489(0.005)} \\
DLGLM & 0.199(0.023) & 0.768(0.130) & 0.698(0.060) & 0.366(0.200) & 0.695(0.235) & 0.480(0.185) \\
MIWAE & 0.203(0.006) & 0.770(0.009) & 0.687(0.013) & 0.359(0.011) & 0.727(0.014) & 0.481(0.008) \\
MICE & 0.195(0.012) & 0.741(0.021) & \textbf{0.708(0.023)} & 0.365(0.015) & 0.612(0.082) & 0.457(0.031) \\
MissForest & 0.197(0.027) & 0.763(0.005) & 0.684(0.061) & 0.363(0.041) & 0.692(0.176) & 0.456(0.062) \\
Mean & 0.221(0.006) & 0.650(0.008) & 0.725(0.014) & 0.233(0.012) & 0.432(0.015) & 0.305(0.007) \\
\bottomrule
\end{tabular}
\end{table}

\begin{table}[htbp]
\centering
\caption{Comparative performance under non-ignorable missingness on NHANES dataset}
\label{tab:nonignorable_comparison}
\scriptsize
\setlength{\tabcolsep}{3pt}
\begin{tabular}{lcccccc}
\toprule
\textbf{Method} & \textbf{Brier Score} & \textbf{AUC} & \textbf{Accuracy} & \textbf{Precision} & \textbf{Recall} & \textbf{F1-Score} \\
\midrule
AV-LR & 0.197(0.006) & \textbf{0.777(0.015)} & 0.684(0.009) & 0.359(0.011) & \textbf{0.742(0.030)} & 0.485(0.016) \\
DLGLM & 0.206(0.031) & 0.770(0.013) & 0.689(0.066) & 0.361(0.010) & 0.720(0.033) & 0.482(0.015) \\
MIWAE & 0.203(0.005) & 0.765(0.011) & 0.686(0.008) & 0.361(0.007) & 0.738(0.004) & \textbf{0.486(0.006)} \\
MICE & \textbf{0.189(0.015)} & 0.759(0.017) & \textbf{0.712(0.030)} & \textbf{0.372(0.017)} & 0.620(0.088) & 0.468(0.017) \\
MissForest & 0.193(0.020) & 0.755(0.027) & 0.692(0.055) & 0.357(0.023) & 0.646(0.195) & 0.452(0.040) \\
Mean & 0.212(0.005) & 0.648(0.010) & 0.712(0.011) & 0.290(0.010) & 0.382(0.019) & 0.301(0.010) \\
\bottomrule
\end{tabular}
\end{table}

\section{Conclusion}

This paper introduced AV‑LR, a unified amortized variational inference framework for binary logistic regression with missing covariates. Unlike existing methods that rely on iterative imputation schemes or introduce complex latent variable structures, AV‑LR performs inference directly in the space of missing data through a simple yet effective architecture. By coupling an amortized inference network with a linear predictive layer and training them jointly via evidence lower bound maximization, AV-LR achieves seamless integration of imputation and classification tasks.
The key advantages of AV-LR are threefold. First, its architectural simplicity, a single inference network and a linear layer, facilitates implementation, optimization, and interpretability while avoiding the computational overhead of nested sampling schemes. Second, the framework naturally extends to non-ignorable missingness (MNAR) through explicit modeling of the missingness mechanism, providing a principled way to handle realistic missing-data patterns that depend on unobserved values. Third, extensive experiments on synthetic and real-world datasets demonstrate that AV-LR attains estimation accuracy comparable to, or better than, state-of-the-art EM-based and deep-generative methods, while offering substantially lower computational cost, especially at inference time.
Looking forward, AV-LR opens several promising research directions. The amortized inference network could be enriched with more flexible variational families (e.g., normalizing flows) to capture complex posteriors, and the framework could be extended to other generalized linear models or to settings with high‑dimensional covariates. Moreover, the principled handling of MNAR mechanisms invites applications in domains where missingness is inherently informative, such as clinical studies or survey data.

\section*{Acknowledgments}
    
\bibliographystyle{elsarticle-num}
\bibliography{biibb}

\newpage
\section*{Appendix A: Comprehensive Experimental Results}

This appendix provides the complete set of experimental results that support the findings in the main text. We present exhaustive performance metrics across all missing data mechanisms, datasets, and evaluation criteria to ensure full transparency and facilitate future comparisons.

\subsection*{A.1 Synthetic Data Experiments}
\label{app:1}
\subsubsection*{A.1.1 MCAR Mechanism (30\% Missingness)}

Table \eqref{tab:appendix_param_rmse_30mcar} presents comprehensive parameter estimation results under the 30\% MCAR mechanism. The results demonstrate AV-LR's consistent performance across all parameter recovery metrics, with particular strength in covariance matrix estimation as evidenced by the lowest Frobenius norm discrepancy.

\begin{table}[htbp]
\centering
\caption{Complete parameter estimation under 30\% MCAR mechanism (mean ± std over 5 repetitions)}
\label{tab:appendix_param_rmse_30mcar}
\begin{tabular}{lcccc}
\toprule
\textbf{Estimator} & \textbf{RMSE\_imp} & \textbf{RMSE\_mu} & \textbf{Frobenius\_cov} & \textbf{RMSE\_beta} \\
\midrule
AV-LR & 0.7145 $\pm$ 0.0078 & 0.00580 $\pm$ 0.00185 & 0.0720 $\pm$ 0.0150 & 0.0445 $\pm$ 0.0130 \\
IDLGLM & 0.7420 $\pm$ 0.0036 & 0.00710 $\pm$ 0.00100 & 0.2450 $\pm$ 0.0120 & 0.1050 $\pm$ 0.0160 \\
MIWAE & 0.7385 $\pm$ 0.0039 & 0.00690 $\pm$ 0.00110 & 0.2180 $\pm$ 0.0090 & 0.0960 $\pm$ 0.0180 \\
SAEM & 0.7320 $\pm$ 0.0025 & 0.00640 $\pm$ 0.00145 & 0.2060 $\pm$ 0.0110 & 0.0780 $\pm$ 0.0200 \\
MICE & 0.7480 $\pm$ 0.0041 & 0.00740 $\pm$ 0.00120 & 0.3150 $\pm$ 0.0180 & 0.1600 $\pm$ 0.0600 \\
MissForest & 0.8025 $\pm$ 0.0030 & 0.00790 $\pm$ 0.00180 & 0.2810 $\pm$ 0.0100 & 0.2850 $\pm$ 0.0550 \\
Mean & 0.9950 $\pm$ 0.0072 & 0.00830 $\pm$ 0.00275 & 1.5200 $\pm$ 0.0150 & 0.4700 $\pm$ 0.0520 \\
\bottomrule
\end{tabular}
\end{table}

The predictive performance under the same experimental conditions is detailed in Table \eqref{tab:appendix_pred_metrics_30mcar}. AV-LR maintains competitive performance across all classification metrics, with particular advantages in AUC and F1-score, indicating balanced precision-recall characteristics.

\begin{table}[htbp]
\centering
\caption{Complete prediction metrics under 30\% MCAR mechanism (mean $\pm$ std over 5 repetitions)}
\label{tab:appendix_pred_metrics_30mcar}
\begin{tabular}{lccccc}
\toprule
\textbf{Estimator} & \textbf{AUC} & \textbf{Accuracy} & \textbf{Precision} & \textbf{Recall} & \textbf{F1 Score} \\
\midrule
AV-LR & 0.8295 $\pm$ 0.0290 & 0.7600 $\pm$ 0.0300 & 0.7370 $\pm$ 0.0180 & 0.7050 $\pm$ 0.1100 & 0.7160 $\pm$ 0.0610 \\
IDLGLM & 0.8210 $\pm$ 0.0260 & 0.7560 $\pm$ 0.0250 & 0.7350 $\pm$ 0.0230 & 0.7000 $\pm$ 0.0780 & 0.7140 $\pm$ 0.0460 \\
MIWAE & 0.8155 $\pm$ 0.0245 & 0.7505 $\pm$ 0.0220 & 0.7230 $\pm$ 0.0230 & 0.6950 $\pm$ 0.0710 & 0.7080 $\pm$ 0.0440 \\
SAEM & 0.8160 $\pm$ 0.0300 & 0.7520 $\pm$ 0.0280 & 0.7260 $\pm$ 0.0290 & 0.6990 $\pm$ 0.0850 & 0.7110 $\pm$ 0.0580 \\
MICE & 0.8130 $\pm$ 0.0310 & 0.7485 $\pm$ 0.0330 & 0.7200 $\pm$ 0.0270 & 0.6900 $\pm$ 0.0880 & 0.7050 $\pm$ 0.0530 \\
MissForest & 0.8020 $\pm$ 0.0315 & 0.7460 $\pm$ 0.0290 & 0.7130 $\pm$ 0.0260 & 0.6870 $\pm$ 0.0830 & 0.7020 $\pm$ 0.0550 \\
Mean & 0.7935 $\pm$ 0.0250 & 0.7380 $\pm$ 0.0240 & 0.7100 $\pm$ 0.0210 & 0.6720 $\pm$ 0.1200 & 0.6880 $\pm$ 0.0700 \\
\bottomrule
\end{tabular}
\end{table}

Computational efficiency represents a critical practical consideration. Table \eqref{tab:appendix_exec_time_30mcar} reveals that AV-LR achieves competitive training times while maintaining minimal inference latency, underscoring its suitability for real-time applications.

\begin{table}[htbp]
\centering
\caption{Execution times under 30\% MCAR mechanism (seconds, mean ± std over 5 repetitions)}
\label{tab:appendix_exec_time_30mcar}
\begin{tabular}{lcc}
\toprule
\textbf{Estimator} & \textbf{Training Time} & \textbf{Testing Time} \\
\midrule
AV-LR & 33.116 $\pm$ 2.649 & 0.019 $\pm$ 0.0015 \\
IDLGLM & 65.229 $\pm$ 7.828 & 6.921 $\pm$ 0.831 \\
MIWAE & 34.425 $\pm$.684 & 1.115 $\pm$ 0.087 \\
SAEM & 3108.24 $\pm$ 134.202 & 28.806 $\pm$ 2.93 \\
MICE & 0.167 $\pm$ 0.015 & 0.100 $\pm$ 0.003 \\
MissForest & 48.490 $\pm$ 3.879 & 25.448 $\pm$ 2.036 \\
Mean & 0.023 $\pm$ 0.001 & 0.004 $\pm$ 0.0002 \\
\bottomrule
\end{tabular}
\end{table}

\subsubsection*{A.1.2 MNAR Mechanism (60\% Missingness)}

The non-ignorable missingness scenario presents the most challenging conditions for all methods. As shown in Table~\ref{tab:appendix_param_rmse_60mnar}, explicitly modeling the missingness mechanism provides AV-LR with a clear advantage under MNAR conditions, particularly in terms of parameter recovery and discriminative performance. The predictive results in Table~\ref{tab:appendix_pred_metrics_60mnar} further demonstrate AV-LR’s robustness, maintaining strong classification performance despite the challenging missingness mechanism. Finally, although modeling non-ignorable mechanisms increases computation time for all methods (see Table~\ref{tab:appendix_exec_time_60mnar}), AV-LR preserves its efficiency advantage, especially during inference.

\begin{table}[htbp]
\centering
\caption{Complete parameter estimation under 60\% MNAR mechanism (mean ± std over 5 repetitions)}
\label{tab:appendix_param_rmse_60mnar}
\begin{tabular}{lcccc}
\toprule
\textbf{Estimator} & \textbf{RMSE\_imp} & \textbf{RMSE\_mu} & \textbf{Frobenius\_cov} & \textbf{RMSE\_beta} \\
\midrule
AV-LR & 0.8624 $\pm$  0.0300 & 0.3300 $\pm$  0.0200 & 1.3000 $\pm$  0.0800 & 0.5400 $\pm$  0.2100 \\
DLGLM & 0.8800 $\pm$  0.0350 & 0.3400 $\pm$  0.0200 & 1.3500 $\pm$  0.0850 & 0.5600 $\pm$  0.2200 \\
NotMIWAE & 0.9200 $\pm$  0.0300 & 0.3600 $\pm$  0.0200 & 1.4200 $\pm$  0.1100 & 0.6000 $\pm$  0.2300 \\
SAEM & 1.3000 $\pm$  0.0200 & 0.6200 $\pm$  0.0250 & 2.0000 $\pm$  0.0800 & 0.6600 $\pm$  0.2900 \\
MICE & 1.2579 $\pm$  0.0132 & 0.5791 $\pm$  0.0036 & 2.0530 $\pm$  0.0849 & 0.5545 $\pm$  0.3647 \\
MissForest & 1.0749 $\pm$  0.0624 & 0.6280 $\pm$  0.0382 & 2.0712 $\pm$  0.1309 & 0.7120 $\pm$  0.3950 \\
Mean & 1.4825 $\pm$  0.0222 & 0.7354 $\pm$  0.0075 & 2.9295 $\pm$  0.0188 & 0.7709 $\pm$  0.3418 \\
\bottomrule
\end{tabular}
\end{table}

\begin{table}[htbp]
\centering
\caption{Complete prediction metrics under 60\% MNAR mechanism (mean ± std over 5 repetitions)}
\label{tab:appendix_pred_metrics_60mnar}
\begin{tabular}{lccccc}
\toprule
\textbf{Estimator} & \textbf{AUC} & \textbf{Accuracy} & \textbf{Precision} & \textbf{Recall} & \textbf{F1 Score} \\
\midrule
AV-LR & 0.771 $\pm$ 0.060 & 0.724 $\pm$ 0.050 & 0.704 $\pm$ 0.070 & 0.704 $\pm$  0.070 & 0.704 $\pm$  0.065 \\
DLGLM & 0.768 $\pm$  0.055 & 0.716 $\pm$  0.045 & 0.694 $\pm$  0.065 & 0.700 $\pm$  0.075 & 0.697 $\pm$  0.060 \\
NotMIWAE & 0.738 $\pm$  0.117 & 0.682 $\pm$  0.097 & 0.657 $\pm$  0.130 & 0.665 $\pm$  0.084 & 0.661 $\pm$  0.060 \\
SAEM & 0.625 $\pm$  0.063 & 0.600 $\pm$  0.045 & 0.630 $\pm$  0.042 & 0.730 $\pm$  0.081 & 0.690 $\pm$ 0.051 \\
MICE & 0.659 $\pm$  0.045 & 0.520 $\pm$  0.051 & 0.489 $\pm$  0.044 & 0.648 $\pm$  0.091 & 0.557 $\pm$  0.057 \\
MissForest & 0.725 $\pm$  0.058 & 0.666 $\pm$  0.060 & 0.615 $\pm$  0.070 & 0.760 $\pm$  0.110 & 0.679 $\pm$  0.080 \\
Mean & 0.634 $\pm$  0.078 & 0.508 $\pm$ 0.054 & 0.479 $\pm$  0.117 & 0.644 $\pm$  0.219 & 0.549 $\pm$  0.098 \\
\bottomrule
\end{tabular}
\end{table}

\begin{table}[htbp]
\centering
\caption{Execution times under 60\% MNAR mechanism (seconds, mean ± std over 5 repetitions)}
\label{tab:appendix_exec_time_60mnar}
\begin{tabular}{lcc}
\toprule
\textbf{Estimator} & \textbf{Training Time} & \textbf{Testing Time} \\
\midrule
AV-LR & 62.915 $\pm$  6.292 & 0.156 $\pm$  0.016 \\
DLGLM & 80.556 $\pm$  8.056 & 6.078 $\pm$  0.608 \\
NotMIWAE & 44.607 $\pm$  4.461 & 1.232 $\pm$  0.123 \\
SAEM & 4211.332 $\pm$  181.877 & 36.433 $\pm$  6.511 \\
MICE & 0.118 $\pm$  0.020 & 0.079 $\pm$  0.010 \\
MissForest & 30.708 $\pm$  5.000 & 14.315 $\pm$  2.500 \\
Mean & 0.012 $\pm$  0.002 & 0.000 $\pm$  0.000 \\
\bottomrule
\end{tabular}
\end{table}

\subsection*{A.2 Real-World Dataset Comprehensive Analysis}
\label{all_REAL}
\begin{figure}[htbp]
  \centering
  \includegraphics[width=\linewidth]{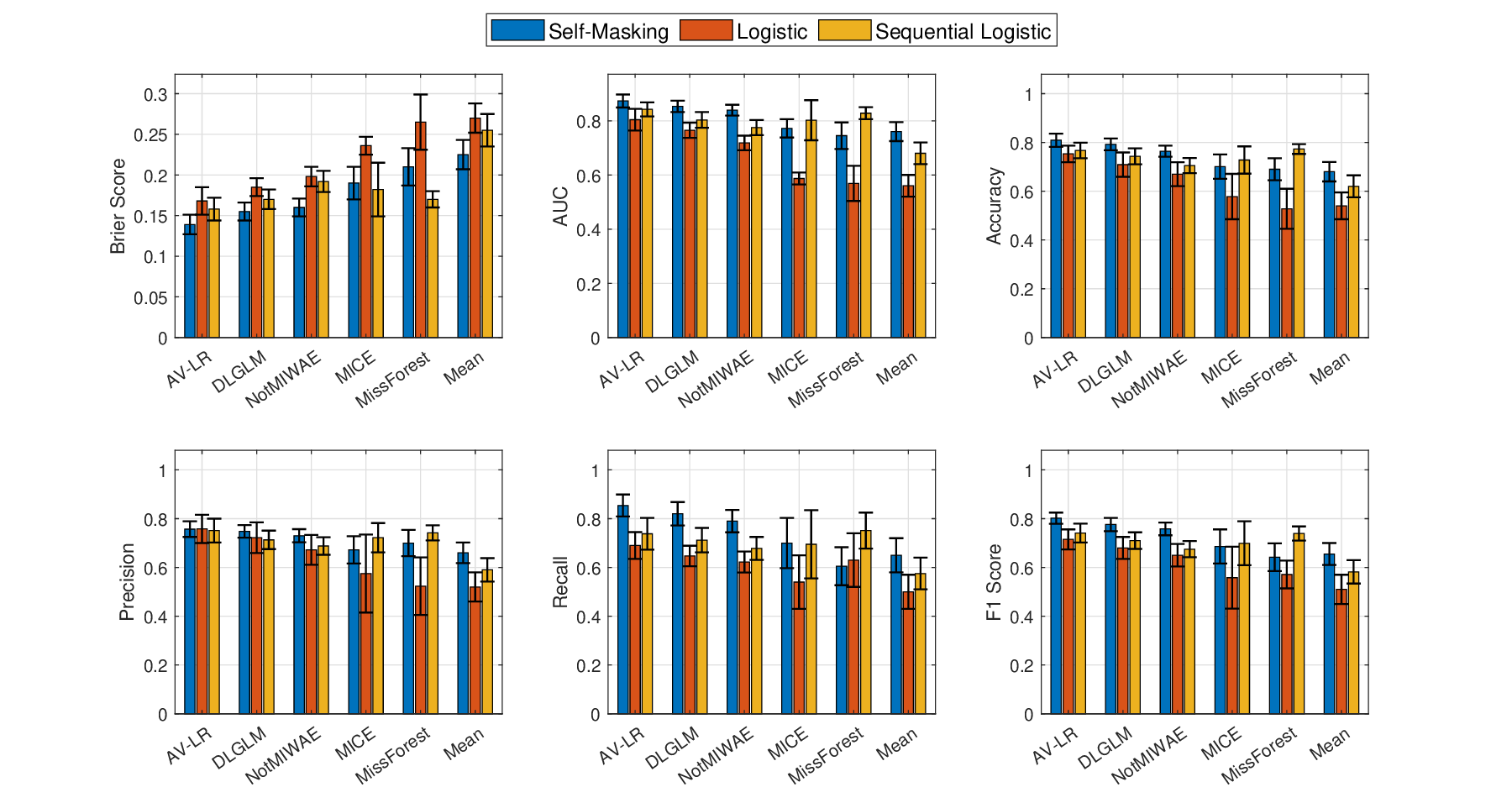}
  \caption{Comparative performance of imputation methods on the Bank Note Authentication Dataset under 50\% MNAR across Self-Masking, Logistic, and Sequential Logistic mechanisms.}

\label{fig:classification_metrics_Bank_Note_Mnar}
\end{figure}

\begin{figure}[htbp]
  \centering
  \includegraphics[width=\linewidth]{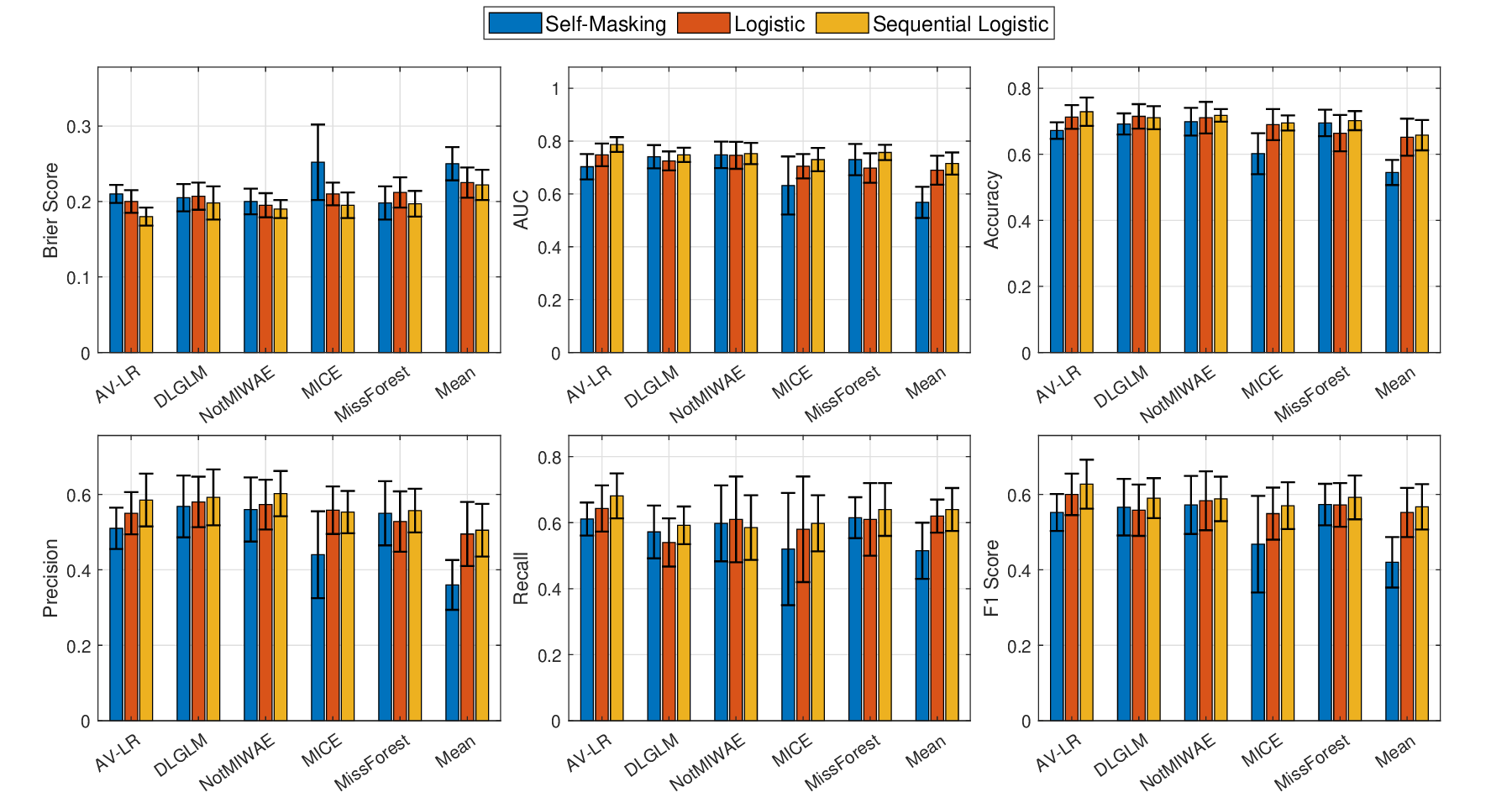}
  \caption{Comparative performance of imputation methods on the Pima Indians Diabetes Database under 50\% MNAR across Self-Masking, Logistic, and Sequential Logistic mechanisms.}

\label{fig:classification_metrics_Indian_Mnar}
\end{figure}

\begin{figure}[htbp]
  \centering
  \includegraphics[width=\linewidth]{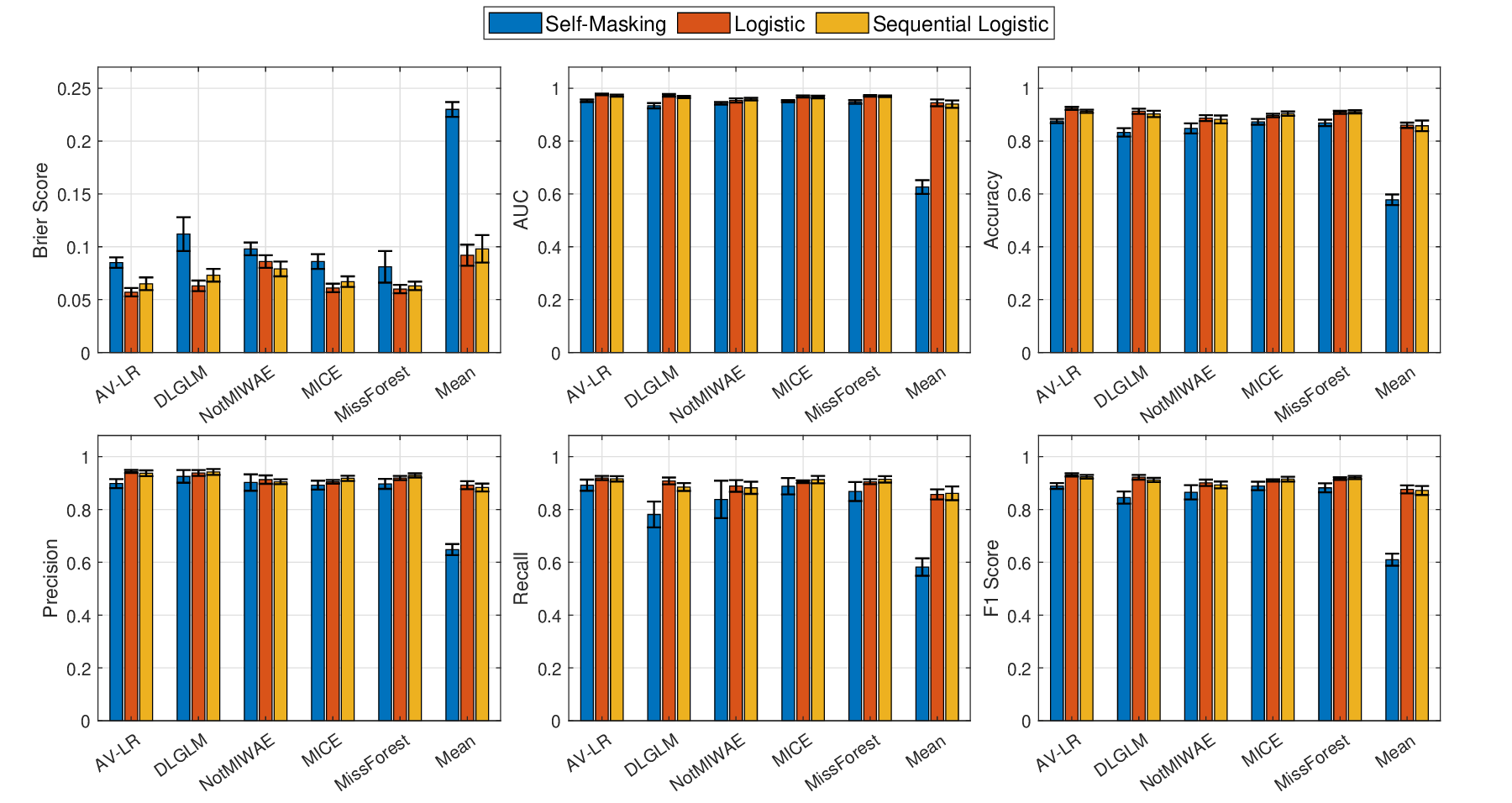}
  \caption{Comparative performance of imputation methods on the Rice Cammeo Osmancik Dataset under 50\% MNAR across Self-Masking, Logistic, and Sequential Logistic mechanisms.}

\label{fig:classification_metrics_Rice_Cammeo_Mnar}
\end{figure}

\begin{figure}[htbp]
  \centering
  \includegraphics[width=\linewidth]{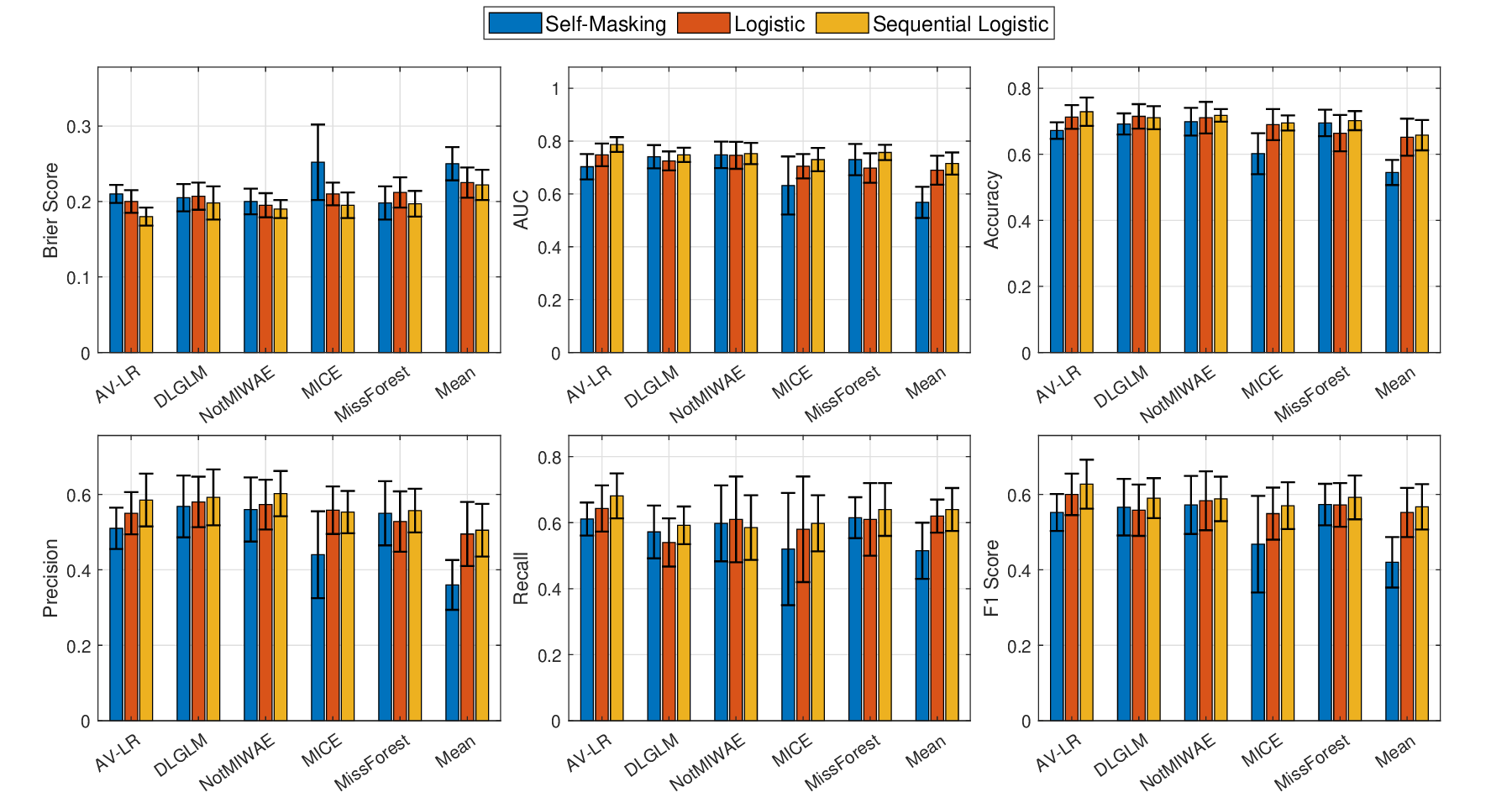}
  \caption{Comparative performance of imputation methods on the Breast Cancer Dataset under 50\% MNAR across Self-Masking, Logistic, and Sequential Logistic mechanisms.}

\label{fig:classification_metrics_Breast_Cancer_Mnar}
\end{figure}

Figures 
\ref{fig:classification_metrics_Bank_Note_Mnar}--\ref{fig:classification_metrics_Breast_Cancer_Mnar} summarize the comparative performance of the evaluated imputation methods (AV-LR, DLGLM, NotMIWAE, MICE, MissForest) across four real-world datasets (Bank Note, Pima, Rice, and Breast Cancer) with 50\% MNAR missingness under three mechanisms (Self-Masking, Logistic, and Sequential Logistic). Each subplot presents the mean scores over multiple replications, with error bars representing standard deviations to convey experimental variability. Overall, AV-LR demonstrates robust and consistent performance across most datasets, while DLGLM and NotMIWAE offer trade-offs between predictive accuracy and variance depending on the dataset and mechanism. MICE is characterized by very low training times but sometimes higher variability, whereas MissForest delivers competitive performance at a substantially higher computational cost on certain datasets. The comprehensive execution time table (Table \ref{tab:execution_times_mnar_no_mean}) highlights these trade-offs, showing contrasting orders of magnitude between lightweight methods (e.g., MICE) and more computationally intensive approaches (e.g., MissForest and certain deep learning variants) in both training and testing phases. Together, these results allow readers to evaluate not only predictive effectiveness but also scalability and practical suitability of each method given computational constraints and data characteristics.

\begin{table*}[htbp]
\centering
\scriptsize
\caption{Execution times (seconds) under 50\% MNAR — Training \& Testing (mean ± std).}
\label{tab:execution_times_mnar_no_mean}
\begin{tabular}{lllcc}
\toprule
\textbf{Dataset} & \textbf{Mechanism} & \textbf{Estimator} & \textbf{Training Time (s)} & \textbf{Testing Time (s)} \\
\midrule
\multicolumn{5}{l}{\textbf{BankNote}} \\
 & Self-Masking & AV-LR      & 45.424 $\pm$  4.540   & 0.0778 $\pm$  0.0143 \\
 &              & DLGLM      & 54.923 $\pm$  5.500   & 4.0530 $\pm$  0.9500 \\
 &              & NotMIWAE   & 13.381 $\pm$  1.380   & 0.5760 $\pm$  0.1100 \\
 &              & MICE       & 0.090 $\pm$  0.021    & 0.0840 $\pm$  0.0245 \\
 &              & MissForest & 18.584 $\pm$ 2.800   & 14.802 $\pm$  1.500 \\
\cmidrule(lr){2-5}
 & Logistic     & AV-LR      & 42.710 $\pm$  3.211   & 0.0756 $\pm$  0.0120 \\
 &              & DLGLM      & 52.333 $\pm$ 4.100   & 4.0606 $\pm$  0.9500 \\
 &              & NotMIWAE   & 13.796 $\pm$  1.900   & 0.5351 $\pm$  0.1100 \\
 &              & MICE       & 0.011 $\pm$  0.003    & 0.0010 $\pm$  0.0005 \\
 &              & MissForest & 16.882 $\pm$  2.100   & 13.8367 $\pm$  1.500 \\
\cmidrule(lr){2-5}
 & Sequential Logistic & AV-LR      & 42.539 $\pm$  4.254   & 0.0760 $\pm$  0.0165 \\
 &                     & DLGLM      & 57.158 $\pm$  5.716   & 5.4339 $\pm$  0.9501 \\
 &                     & NotMIWAE   & 13.875 $\pm$  1.388   & 0.5287 $\pm$  0.1130 \\
 &                     & MICE       & 0.086 $\pm$  0.021    & 0.0767 $\pm$  0.0222 \\
 &                     & MissForest & 17.663 $\pm$  2.650   & 13.742 $\pm$  1.512 \\
\midrule
\multicolumn{5}{l}{\textbf{Pima}} \\
 & Self-Masking & AV-LR      & 22.260 $\pm$  2.226   & 0.06249 $\pm$  0.0100 \\
 &              & DLGLM      & 26.409 $\pm$  2.641   & 3.19393 $\pm$  0.9500 \\
 &              & NotMIWAE   & 20.421 $\pm$  1.800   & 0.28680 $\pm$  0.0800 \\
 &              & MICE       & 0.147 $\pm$  0.030    & 0.13068 $\pm$  0.0300 \\
 &              & MissForest & 31.859 $\pm$  3.186   & 25.356 $\pm$  2.536 \\
\cmidrule(lr){2-5}
 & Logistic     & AV-LR      & 25.261 $\pm$  3.000   & 0.0625 $\pm$ 0.0100 \\
 &              & DLGLM      & 29.710 $\pm$  4.000   & 3.3973 $\pm$  0.9500 \\
 &              & NotMIWAE   & 25.648 $\pm$  1.800   & 0.3321 $\pm$  0.0800 \\
 &              & MICE       & 0.167 $\pm$  0.020    & 0.0907 $\pm$  0.0100 \\
 &              & MissForest & 36.747 $\pm$  4.500   & 28.859 $\pm$  3.000 \\
\cmidrule(lr){2-5}
 & Sequential Logistic & AV-LR      & 23.269 $\pm$  2.327   & 0.06709 $\pm$  0.0105 \\
 &                     & DLGLM      & 26.196 $\pm$  2.620   & 3.31089 $\pm$  0.9500 \\
 &                     & NotMIWAE   & 21.390 $\pm$  2.139   & 0.49171 $\pm$  0.0800 \\
 &                     & MICE       & 0.169 $\pm$  0.030    & 0.14863 $\pm$  0.0300 \\
 &                     & MissForest & 33.688 $\pm$  3.369   & 26.669 $\pm$ 2.667 \\
\midrule
\multicolumn{5}{l}{\textbf{Rice}} \\
 & Self-Masking & AV-LR      & 105.042 $\pm$  10.504 & 0.26876 $\pm$  0.0422 \\
 &              & DLGLM      & 123.146 $\pm$  12.315 & 8.74180 $\pm$  1.266 \\
 &              & NotMIWAE   & 28.510 $\pm$  2.851   & 1.60855 $\pm$  0.1610 \\
 &              & MICE       & 0.171 $\pm$  0.031    & 0.12817 $\pm$  0.0302 \\
 &              & MissForest & 60.943 $\pm$  6.094   & 29.888 $\pm$  3.022 \\
\cmidrule(lr){2-5}
 & Logistic     & AV-LR      & 106.781 $\pm$  10.678 & 0.3910 $\pm$  0.047 \\
 &              & DLGLM      & 122.716 $\pm$  12.272 & 8.5769 $\pm$ 1.200 \\
 &              & NotMIWAE   & 28.511 $\pm$  2.851   & 1.5412 $\pm$  0.200 \\
 &              & MICE       & 0.332 $\pm$  0.080    & 0.1311 $\pm$  0.030 \\
 &              & MissForest & 63.448 $\pm$  6.345   & 32.696 $\pm$  3.300 \\
\cmidrule(lr){2-5}
 & Sequential Logistic & AV-LR      & 111.530 $\pm$  11.153 & 0.37294 $\pm$  0.0411 \\
 &                     & DLGLM      & 154.098 $\pm$  15.410 & 10.6019 $\pm$  1.060 \\
 &                     & NotMIWAE   & 31.293 $\pm$  3.129   & 1.75548 $\pm$  0.176 \\
 &                     & MICE       & 0.390 $\pm$  0.078    & 0.20573 $\pm$  0.041 \\
 &                     & MissForest & 81.410 $\pm$  8.141   & 31.594 $\pm$  3.159 \\
\midrule
\multicolumn{5}{l}{\textbf{Breast Cancer}} \\
 & Self-Masking & AV-LR      & 20.902 $\pm$  2.090   & 0.08677 $\pm$  0.011 \\
 &              & DLGLM      & 50.345 $\pm$  5.035   & 3.94796 $\pm$  0.395 \\
 &              & NotMIWAE   & 70.676 $\pm$  7.068   & 0.22013 $\pm$  0.030 \\
 &              & MICE       & 0.943 $\pm$ 0.094    & 0.58582 $\pm$  0.060 \\
 &              & MissForest & 156.476 $\pm$  15.648 & 112.416 $\pm$  11.242 \\
\cmidrule(lr){2-5}
 & Logistic     & AV-LR      & 18.051 $\pm$  1.805   & 0.08274 $\pm$  0.010 \\
 &              & DLGLM      & 34.266 $\pm$  3.427   & 5.37148 $\pm$  0.537 \\
 &              & NotMIWAE   & 65.371 $\pm$  6.537   & 0.37380 $\pm$  0.075 \\
 &              & MICE       & 0.682 $\pm$  0.068    & 0.65465 $\pm$  0.066 \\
 &              & MissForest & 203.961 $\pm$  20.396 & 127.160 $\pm$  12.716 \\
\cmidrule(lr){2-5}
 & Sequential Logistic & AV-LR      & 19.254 $\pm$  1.925   & 0.04987 $\pm$  0.005 \\
 &                     & DLGLM      & 47.571 $\pm$  4.757   & 4.17038 $\pm$  0.417 \\
 &                     & NotMIWAE   & 65.930 $\pm$  6.593   & 0.29920 $\pm$  0.030 \\
 &                     & MICE       & 1.152 $\pm$  0.230    & 1.499 $\pm$  0.300 \\
 &                     & MissForest & 168.923 $\pm$  16.892 & 134.825 $\pm$ 13.482 \\
\bottomrule
\end{tabular}
\end{table*}





\end{document}